\PassOptionsToPackage{table}{xcolor}
\documentclass[10pt]{article}

\usepackage{microtype}
\usepackage{graphicx}
\usepackage{subfigure}
\usepackage{booktabs} 
\usepackage[T1]{fontenc}
\usepackage{courier}
\usepackage{listings}
\usepackage{enumitem}

\usepackage[most]{tcolorbox}

\usepackage{hyperref}
\usepackage{multicol}
\usepackage{multirow}
\usepackage{wrapfig}

\definecolor{best}{HTML}{C6E0B4}
\definecolor{secondbest}{HTML}{E2F0D9}
\definecolor{worst}{HTML}{F4C2C2}
\definecolor{secondworst}{HTML}{F9E0E0}

\definecolor{best1}{HTML}{B2DFDB}
\definecolor{best2}{HTML}{CDEAE8}
\definecolor{best3}{HTML}{E0F2F1}
\definecolor{best4}{HTML}{F1F8F7}
\definecolor{worst1}{HTML}{FFCDD2}
\definecolor{worst2}{HTML}{F8E1E3}
\definecolor{worst3}{HTML}{FCEBED}
\definecolor{worst4}{HTML}{FFF8F9}

\tcbset{
  colback=best3,     
  colframe=best1,    
  boxrule=0.4pt,
  arc=2pt,
  left=6pt,right=6pt,top=5pt,bottom=5pt,
}

\newtcolorbox{promptbox}[1][]{
  enhanced,
  breakable,
  #1
}

\usepackage{arxiv}
\usepackage{natbib}

\renewcommand{\headeright}{}
\renewcommand{\undertitle}{}
\makeatletter
\renewcommand{\@toptitlebar}{\vskip 0.25in\vskip -\parskip}
\renewcommand{\@bottomtitlebar}{\vskip 0.29in\vskip -\parskip\vskip 0.09in}
\makeatother

\usepackage{setspace}
\definecolor{hsbdeep}{HTML}{1E6669}   
\definecolor{hsbfg}{HTML}{16292B}     
\definecolor{hsbbg}{HTML}{EDF4F3}     
\definecolor{hsbrule}{HTML}{B2DFDB}   

\newcommand{\hsbaffiliations}{}
\newcommand{\hsbcorrespondence}{}
\newcommand{\abstracttext}{}
\newcommand{\afmark}[1]{\textsuperscript{\normalfont\color{hsbdeep}#1}}
\newcommand{\email}[1]{\href{mailto:#1}{\texttt{#1}}}

\makeatletter
\renewcommand{\@maketitle}{%
  \vbox{%
    \hsize\textwidth
    \linewidth\hsize
    \begin{tcolorbox}[
      enhanced, frame hidden, breakable=false,
      colback=hsbbg, colframe=hsbbg, boxrule=0pt,
      arc=10pt, outer arc=10pt,
      left=0.55cm, right=0.55cm, top=0.55cm, bottom=0.4cm,
      grow to left by=1.5pt, grow to right by=1.5pt,
      before skip=0pt,
      overlay={\node[anchor=south east, at=(frame.south east),
        xshift=-0.55cm, yshift=0.42cm,
        font=\footnotesize\sffamily\bfseries, text=hsbdeep]
        {\MakeUppercase{Technical Report}};}
    ]
    \setlength{\parindent}{0pt}%
    \color{hsbfg}%
    {\raggedright
      {\setstretch{1.25}\huge\sffamily\bfseries\color{hsbdeep}\@title\par}%
      \vskip 0.34cm
      {\sffamily\bfseries\@author\par}%
      \vskip 0.24cm
      {\normalsize\hsbaffiliations\par}%
    }%
    \vskip 0.40cm
    {\abstracttext\par}%
    \vskip 0.26cm
    {\small{\sffamily\bfseries\color{hsbdeep}Correspondence:} \hsbcorrespondence\par}%
    \end{tcolorbox}
  }
}
\makeatother

\usepackage[utf8]{inputenc} 
\usepackage[T1]{fontenc}    
\usepackage{hyperref}       
\usepackage{url}            
\usepackage{booktabs}       
\usepackage{amsfonts}       
\usepackage{nicefrac}       
\usepackage{microtype}      

\usepackage{amsmath}
\usepackage{amssymb}
\usepackage{mathtools}
\usepackage{amsthm}

\usepackage[capitalize,noabbrev]{cleveref}
\usepackage{tabularx} 
\theoremstyle{plain}

\theoremstyle{definition}

\theoremstyle{remark}

\usepackage[textsize=tiny]{todonotes}
\usepackage{bbm}
\usepackage{siunitx}

\title{HumanStudy-Bench: Towards AI Agent Design for Participant Simulation}
\date{}  

\author{%
Xuan Liu\afmark{1,6}, HaoYang Shang\afmark{6} \\[0.14cm]
Yuanjun Feng\afmark{2,6}, Zizhang Liu\afmark{3,6}, Xinyan Liu\afmark{6} \\[0.14cm]
Guankai Zhai\afmark{4}, Yunze Xiao\afmark{5}, Yiwen Tu\afmark{1}, Haojian Jin\afmark{1}%
}

\renewcommand{\hsbaffiliations}{%
\afmark{1}University of California, San Diego, \afmark{2}Universit\'e de Lausanne,
\afmark{3}Tsinghua University, \afmark{4}Stanford University,
\afmark{5}Carnegie Mellon University, \afmark{6}HumanStudy Hub%
}

\renewcommand{\hsbcorrespondence}{Xuan Liu at \email{xul049@ucsd.edu}}

\begin{document}

\renewcommand{\abstracttext}{%

Large language models (LLMs) are increasingly used as simulated participants in social science experiments, but their behavior is often unstable and highly sensitive to design choices. Prior evaluations frequently conflate base model capabilities with experimental instantiation, obscuring whether outcomes reflect the model itself or the agent setup. We instead frame \textbf{participant simulation as an agent-design problem} over full experimental protocols, where an agent is defined by a base model and a specification (e.g., participant attributes) that encodes behavioral assumptions. We introduce \textsc{HUMANSTUDY-BENCH}, an open platform and execution engine designed for practitioners to develop and evaluate agents tailored to their target experimental settings. The platform reconstructs published human-subject experiments via a human-in-the-loop Filter--Extract--Execute--Evaluate pipeline that preserves the original stimuli, conditions, and statistical procedures end to end, while allowing practitioners to freely explore the agent design space. We introduce two complementary metrics that quantify agreement with humans on both the significance conclusion and the effect size, while accounting for finite-sample uncertainty in the human reference data. In collaboration with social scientists, we validate the platform on a suite of 12 foundational studies covering 6,000+ trials across individual cognition, strategic interaction, and social psychology. 

}

\maketitle

\section{Introduction}
\label{sec:intro}
Recent work shows that large language models (LLMs) exhibit human-like decision-making and social behavior across diverse domains, including economic experiments~\citep{horton2023large_eco}, political science~\citep{argyle2023out, Hofmann2024_nature}, marketing~\citep{li2024frontiers}, and social psychology~\citep{dillion2023can, aher2023using}. These findings have motivated the use of LLM-based models as surrogates or simulation testbeds for human participants in social science research~\citep{gao2024large, manning2024automated, position_CHI, anthis2025position}. However, converging evidence cautions against treating such LLM-based models as faithful simulators of human subjects~\citep{participants_nature, doi:10.1073/pnas.2501660122}. Across a range of behavioral tasks, they often yield unstable response distributions~\citep{pmlr-v202-santurkar23a}, lack sensitivity to population heterogeneity~\citep{Bisbee_Clinton_Dorff_Kenkel_Larson_2024}, and are overly sensitive to experimental design choices like prompt wording or sampling settings~\citep{loya-etal-2023-exploring, sclar2024quantifying}.

These limitations pose a fundamental challenge for using raw models as ``drop-in'' replacements for human participants. For social science applications, the key requirement is not only plausible individual responses, but also stable condition-level effects and reliable inferential conclusions under the same analysis pipeline used for human data~\citep{PNAS_Direct, ying2025benchmarkinghuman}. Yet most existing evaluations either characterize raw model behavior on bespoke tasks~\citep{Measvalue, hu2025simbench}, or test whether a fixed prompt can qualitatively reproduce canonical effects~\citep{exploring_iclr2025}. Such setups conflate base model capability with experimental instantiation, obscuring the iterative design work that largely determines whether a simulator is usable in practice.

In practice, LLMs enter social science workflows as surrogate participant \emph{agents}: depending on the application, a base model may be instantiated with explicit roles and task framing, and optionally equipped with demographic attributes, goals, memory, or auxiliary mechanisms (e.g., tools or skills). Such an agent is typically embedded into a full experimental protocol. Crucially, these specifications are not neutral wrappers—they encode behavioral assumptions and can qualitatively change outcomes even when the base model is held fixed. This motivates treating participant simulation as an \emph{agent design} problem under realistic constraints (e.g., a limited number of design iterations), and, in turn, evaluating alignment at the level of scientific inference across heterogeneous paradigms. 

To this end, we introduce \textsc{HumanStudy-Bench}, an open platform and execution engine that decouples experimental setup from agent design, allowing practitioners to develop and evaluate agents tailored to their target experimental settings. First, it is explicitly developed to operationalize \emph{agent design}: each agent comprises a base model and a specification (e.g., roles, demographics, prior experience, or auxiliary mechanisms) that instantiates behavioral hypotheses,  encoding assumptions about which aspects of human participants matter for reproducing the original effect.

Second, \textsc{HumanStudy-Bench} provides a high-fidelity execution engine that reconstructs full human-subject experiments from published studies and replays their stimuli, instructions, and trial sequences under the original experimental conditions and statistical procedures via an end-to-end human-in-the-loop Filter--Extract--Execute--Evaluate pipeline. This turns static articles into a reusable simulation environment that provides a controlled testbed for comparing different \textit{agent designs} under identical experimental conditions and analysis pipelines.

Finally, we introduce two complementary metrics that score the two facets of any hypothesis test. The \emph{Probability Alignment Score} (PAS) measures \emph{inferential} agreement: whether agents and humans reach the same significance conclusion about an effect. The \emph{Effect Consistency Score} (ECS) measures \emph{magnitude} agreement: whether agents also reproduce the size of the human effect. The two can diverge; an agent may reach the right conclusion while exaggerating the effect, so we report both.

Throughout, agents generate individual trial-level responses that are analyzed with the study's original statistical pipeline, rather than being asked directly what the study concluded. Many of these conclusions are population-level constructs---subgroup comparisons, interaction effects, variance-based statistics, and path-dependent multi-round dynamics---that exist only through aggregation over heterogeneous individuals and cannot be recovered from a single oracle call.

As an initial validation of the platform, we curated a suite of 12 foundational studies from around 100 candidates in collaboration with social scientists, covering 6,000+ trials with human samples ranging from tens to over 2,100 participants. Using \textsc{HumanStudy-Bench}, we evaluate 10 contemporary LLMs and four common agent designs. Our results show that current LLM-based agents achieve limited and inconsistent inferential alignment with humans: they exhibit polarized, bimodal behaviors rather than human-like unimodal patterns, agent design has a large and non-monotonic effect, and neither larger models nor simple multi-model ensembles reliably improve alignment.





\section{Background \& Related Work}
\textbf{LLM-based social simulation.}
LLM-based social simulation uses LLMs as simulated human participants to study social and behavioral phenomena~\citep{horton2023large_eco, argyle2023out, gao2024large, manning2024automated}. 
Recent work also uses multimodal language models such as VLMs for settings with visual or other non-textual stimuli~\citep{huang2026probingmultimodallargelanguage}.
Most work constructs agents via prompting, including \emph{persona prompting} based on demographic profiles or backstories to mimic specific subpopulations~\citep{liu2024sophia, chen2024agentverse, jiang-etal-2024-personallm} and \emph{context-rich prompting} that augments agents with richer evidence such as comprehensive memory streams~\citep{park2023generativeagentsinteractivesimulacra, park2024generativeagentsimulations1000}. 
There are also representation-level methods such as \emph{steering vectors}, which inject latent features for fine-grained control~\citep{Cobra}, and \emph{fine-tuning}, which directly aligns models with human behavior~\citep{SociallyAligned2024, kolluri2025finetuningllmshumanbehavior, binz2025foundation}. 

\textbf{Evaluations.}
The growth of LLM-based social simulation creates a need for systematic evaluation.
Existing benchmarks typically evaluate models on static datasets of survey questions or bespoke tasks~\citep{dominguez-olmedo2024questioning, Measvalue, hu2025simbench}, for example, assessing whether models reproduce canonical social science findings~\citep{exploring_iclr2025} or match human responses on standard psychometric inventories~\citep{miotto-etal-2022-gpt, jiang2023evaluating, serapio2025psychometric}. 
On the task side, they typically treat LLMs as raw models in single-turn or fixed-prompt settings, rather than as configurable agents in realistic multi-stage experiments, providing little guidance on how to design agents for participant simulation in social science research. On the evaluation side, current approaches rely on aggregate point estimates~\citep{argyle2023out, park2024generativeagentsimulations1000} or distributional distance metrics~\citep{suh2025rethinkingllmhumansimulation}; both ignore randomness in human responses and differences in statistical power, and lack theoretical guarantees~\citep{hullman2026human}.


\textsc{HumanStudy-Bench} addresses both gaps: it decouples experimental setup from agent design, allowing practitioners to explore and optimize agent design against the same reconstructed experimental protocol; it replaces threshold-based significance matching with inference-level metrics that account for finite-sample uncertainty and are comparable across heterogeneous studies.

\section{\textsc{HumanStudy-Bench}}
\begin{figure*}[htbp] 
    \centering
    \includegraphics[width=0.96\textwidth]{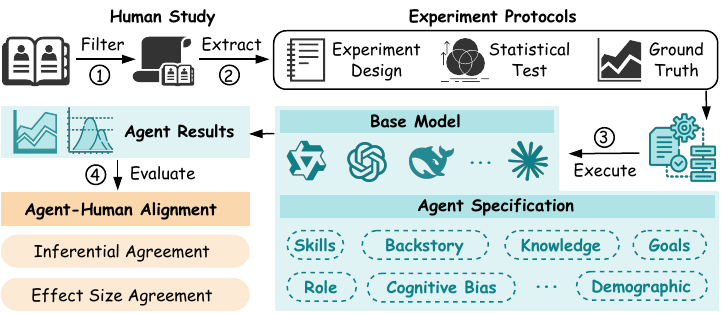}
    \caption{\textbf{Overview of the \textsc{HumanStudy-Bench} engine.} Given published human-subject studies, the engine filters and extracts experimental designs, statistical tests, and human ground-truth results into a reusable simulation environment. Practitioners design agents---combining a base model with a specification (e.g., demographic, backstory, role)---and run them through reconstructed experimental protocols. Agent results are then evaluated against human ground truth via two complementary metrics: inferential agreement (PAS) and effect size agreement (ECS).} 
    \label{fig:overview} 
\end{figure*}

\textsc{HumanStudy-Bench} evaluates \textbf{agent design}: how models are instantiated as AI agents that serve as surrogate participants in social science experiments. Rather than on ``raw'' model capability, we focus on the agent design space, which includes both the base model and the agent specification. Our premise is that agent specifications instantiate behavioral hypotheses, encoding assumptions about which aspects of human participants (e.g., demographics, cognitive biases, task understanding) matter for reproducing the original effect. As a result, AI agents with the same base model can exhibit qualitatively different behavioral patterns under different agent specifications.

\subsection{Task Formulation}
\label{subsec:task-formulation}
\textbf{Study Input.}
To instantiate a simulation, the required input is a human-subject study, including (1) an experimental design
$
\mathcal{E} = (\textit{conditions}, \textit{stimuli}, \textit{measures}, \textit{...}), 
$
(2) a pre-specified statistical hypothesis test (e.g., $H_0: \mu_1 = \mu_2$), and (3) ground-truth human results $D_h$ derived from the original human data (e.g., test statistic, effect direction, and significance decision).

Participants' profiles (e.g., demographics) are optional. They can serve as reference information for the practitioner when designing agents, but are not required for experiment reconstruction.

\textbf{Agent Design.}
Given this study input, the practitioner's task is to design AI agents:
\[
Agent = (\textit{Model}, \textit{Specification}),
\]
with the goal of faithfully reproducing the inferential conclusions of the original human-subject study.
The \textit{Specification} determines how the base \textit{Model} (e.g., GPT, Claude, Gemini) is instantiated as a participant in the experiment, including task and role prompts, participant-facing attributes (e.g., demographics, prior knowledge, goals), and auxiliary mechanisms such as tools, skills, or memory. The agent is then run under the experimental design $\mathcal{E}$ to produce agent ``responses'':
\[
X_a = \{x_1, \ldots, x_{n}\}.
\]

\textbf{Evaluation.}
We evaluate whether agent responses $X_a$ and human responses $X_h$ lead to the same inferential conclusions when passed through the study's original analysis pipeline. We assess this alignment along two complementary axes of any hypothesis test. At the \emph{inferential level}, we ask whether agents and humans agree on whether the effect exists; we capture this with the \textit{Probability Alignment Score} (PAS). At the \emph{magnitude level}, we ask whether agents reproduce the human effect size; we capture this with the \textit{Effect Consistency Score} (ECS). Both metrics are explained in Section~\ref{subsubsec:evaluate}.

\subsection{HumanStudy-Bench Engine}
\label{subsec:Pipeline}
\textsc{HumanStudy-Bench} transitions raw literature into reusable simulation environments through an LLM-assisted, human-in-the-loop pipeline---\textbf{Filter}, \textbf{Extract}, \textbf{Execute}, and \textbf{Evaluate}. The pipeline was co-designed with social scientists through iterative refinement. To ensure scientific rigor, in \textit{Filter} and \textit{Extract} stages, LLM outputs undergo human verification before entering the pipeline; in \textit{Evaluate} stage, all statistical analyses are performed deterministically via standard scientific libraries, with no LLM involvement.

\subsubsection{Filter} 
\label{subsubsec:filter}

To ensure experimental fidelity, the filter stage curates human studies that are both scientifically important and practically reproducible. We include a study only if (i) the full experimental details are documented (e.g., stimuli, task instructions, and experimental conditions; participant profiles are not required); (ii) outcomes are quantifiable with clearly specified statistical tests and reported effect sizes; (iii) the design is simulation-feasible, excluding studies that require specialized equipment or physiological measurements (e.g., fMRI or longitudinal training protocols). These criteria were co-designed with social scientists to balance scientific rigor with simulation feasibility.

We operationalize these criteria using an LLM-assisted filter. Given a candidate research article, the filter parses the paper, extracts experimental details, and assesses the study against the inclusion criteria. It produces a structured verification checklist that flags potential exclusion reasons (e.g., reliance on specialized equipment, missing statistical details). Human reviewers inspect this checklist to verify extraction accuracy, correct any errors, and make the final inclusion decision. Practitioners can apply the same standardized process to curate their own studies.

The study suite used to validate the platform comprises 12 foundational studies, curated from around 100 candidates in collaboration with social scientists (Appendix~\ref{app:study}). We deliberately anchor the initial suite in classic studies drawn from high-impact journals, whose effects are robustly established through independent replication, providing a stable ground truth for evaluating agent design.

\subsubsection{Extraction} 
\label{subsubsec:extract}
The extraction stage formalizes unstructured social science studies into machine-executable representations. We extract the \textit{Experiment Design} (e.g., experimental conditions, factorial structures, trial sequences, and stimulus materials) and the original \textit{Statistical Tests} (e.g., stated hypotheses, and test types) along with human \textit{Ground-Truth} outcomes (e.g., test statistics, $p$-values, and descriptive summaries), which define the evaluation targets for alignment. \textit{Participants' Profiles} (e.g., sample size, demographics, and group assignments) are also extracted when available, serving as optional reference priors for agent specification.

We implement this stage using an LLM-assisted extractor, which parses the filtered papers into a standardized, machine-readable schema. As with filtering, the extractor’s outputs include a verification checklist for each study, which human reviewers inspect to correct errors and approve the extracted materials.

\subsubsection{Execution} 
\label{subsubsec:execute}

\textsc{HumanStudy-Bench} takes a given set of agent designs and runs them through the extracted experimental protocols via a shared execution engine. The key design principle is a task-centric interface that decouples agent implementations from individual studies: different agent designs can be plugged into the same tasks without changing the underlying experimental scripts, enabling controlled comparisons across models, prompts, or other variants of agent specifications.

Each study yields its own configuration module that encodes study-specific procedures---conditions, stimulus order, and instruction prompts. These modules are generated by an LLM-assisted configurator that synthesizes study-specific code from the extracted structured representations. Since the configuration code follows a fixed template, the LLM fills in study-specific details already extracted from the extraction stage rather than generating code from scratch, ensuring reliability while reducing manual effort. For every trial, the execution module packages the relevant inputs into a task instance and feeds it to the agent, which returns a trial-level response (e.g., choices, free-form texts, or ratings) recorded together with trial metadata. All configuration code is cross-reviewed by human researchers before deployment.
\subsection{Measuring Agent--Human Alignment through Validated Hypotheses}
\label{subsubsec:evaluate}

With protocols reconstructed and agent responses collected, the evaluation stage judges alignment at the level of scientific inference: whether agent and human data yield the same inferential conclusions under identical statistical pipelines. Each study has its original analyses reproduced on agent data, and the resulting test statistics are passed to a shared scoring module defined by the PAS and ECS metrics below.

A study typically reports multiple \textit{findings}---distinct behavioral phenomena (e.g., ``Framing Effect''), each supported by one or more \textit{statistical tests}. We score each finding on two complementary axes: the \textit{Probability Alignment Score} (PAS) captures whether agents and humans reach the same inferential conclusion about whether an effect exists; the \textit{Effect Consistency Score} (ECS) captures whether the effect sizes also match. The two metrics are related but not redundant: a study can yield a statistically significant but small effect, or a large but non-significant effect due to insufficient power, so a model can succeed on one axis while failing on the other. Scores are aggregated from tests to findings to studies in a study-balanced manner, so no single study dominates the benchmark.\\

\textbf{Metric 1: Probability Alignment Score (PAS).}
Consider a statistical hypothesis test $j$. Let $\theta_{h,j}, \theta_{a,j} \in \{0, 1\}$ denote whether the effect truly exists ($H_1$) or not ($H_0$) in the human population and the agent simulation. Ideally, we seek the \textit{Oracle Alignment Score}: the true agreement between these latent states,
\begin{equation}
    \mathcal{A}^*_j = \mathbbm{1}(\theta_{a,j} = \theta_{h,j}).
\end{equation}
In practice, the latent truth $\theta$ is never directly observable; we only see noisy data. A naive approach compares binary significance decisions---but this is unstable: studies at $p{=}0.049$ and $p{=}0.051$ receive opposite labels despite near-identical evidence, and finite human samples contain intrinsic uncertainty. PAS instead provides a continuous approximation to $\mathcal{A}^*_j$, following the framework of \textit{Probability of Agreement}~\citep{gwet2014handbook}: it quantifies the probability that agent and human data support the same hypothesis.

The construction follows a standard Bayesian evidence-to-probability pipeline~\citep{bishop2006pattern,jaynes2003probability}. First, each test statistic is converted into a likelihood ratio $\Lambda_j = P(\text{Data}_j \mid H_1) / P(\text{Data}_j \mid H_0)$, measuring the relative support of the data for the alternative versus the null, using established conversions tailored to the test type (see Appendix~\ref{app:prior-choice} for formulas and prior choices). Under the neutral priors adopted below, $\Lambda$ coincides with the Bayes Factor $BF_{10}$. Next, $\Lambda \in [0, \infty)$ is mapped to a bounded posterior probability via the logistic transform:
\begin{equation}
    \pi_{h} = \frac{\Lambda_{h}}{1 + \Lambda_{h}}, \quad \pi_{a} = \frac{\Lambda_{a}}{1 + \Lambda_{a}}
\end{equation}
Under a uniform prior on $\theta$ (\textit{Principle of Indifference}), $\pi$ corresponds to the posterior $P(H_1 \mid \text{Data})$~\citep{bishop2006pattern,jaynes2003probability} (see Appendix~\ref{app:theory} for derivation and frequentist interpretation). Finally, alignment is the probability that human and agent inferences support the same hypothesis---either both $H_1$ or both $H_0$:
\begin{equation}
    \mathcal{A}_j = \underbrace{\pi_{h}\pi_{a}}_{P(\text{Both } H_1)} + \underbrace{(1 - \pi_{h})(1 - \pi_{a})}_{P(\text{Both } H_0)}
\end{equation}
$\mathcal{A}_j \in [0, 1]$: when human evidence is ambiguous ($\pi_h \approx 0.5$), the score converges to $0.5$ rather than penalizing the agent for failing to replicate noise. This property naturally down-weights underpowered tests in aggregation, so that benchmark scores reflect genuine alignment rather than sensitivity to sampling variability. Extension to multiple hypotheses is detailed in Appendix~\ref{app:implementation}. Because an agent may also produce a significant effect in the direction \emph{opposite} to the human one, our implementation scores a three-way outcome space $\{H_{+}, H_{-}, H_{0}\}$ rather than the two-state form above, so that a reversed effect scores near zero (details in Appendix~\ref{app:generalization}).

\textit{Worked example.} In \textit{False Consensus} study, human data yields $F(1, 312) = 49.1$. Under a standard objective prior~\citep{rouder2009bayesian}, this gives $\Lambda_h \approx 10^8$ and $\pi_h \approx 1.0$ (near-certain that the effect exists). If an agent produces a non-significant $F = 1.2$, then $\Lambda_a \approx 0.15$ and $\pi_a \approx 0.13$, yielding $\mathcal{A}_j \approx 0.20$---the agent and humans disagree on whether the effect exists, and PAS reflects this as low alignment.

\paragraph{Aggregation.}
Within each finding, we combine $M$ tests by averaging the alignment scores on the logit scale via the $\operatorname{arctanh}$ transform:
\begin{equation}
\bar{\mathcal{A}}_{\text{finding}} = \frac{1}{2} \left( \tanh \left( \frac{1}{M} \sum_{j=1}^{M} \operatorname{arctanh}(2\mathcal{A}_j - 1) \right) + 1 \right)
\end{equation}
Finding-level scores are then averaged within each study, and the benchmark PAS is the mean over studies, with each study contributing equal weight regardless of how many findings it contains (details in Appendix~\ref{app:aggregation_impl}).\\

\textbf{Metric 2: Effect Consistency Score (ECS).}
PAS measures agreement on whether an effect exists; ECS measures agreement on the effect size. The two capture complementary aspects of replication fidelity: an agent that reproduces a ``framing effect'' but exaggerates it tenfold can score well on PAS yet poorly on ECS, and vice versa.

We adopt a psychometric approach~\citep{campbell1959convergent}, measuring \textit{Concurrent Validity} of the agent's behaviors using Standardized Effect Sizes (Appendix~\ref{app:effect_size}). The score has to catch two failure modes simultaneously: an agent that gets the \emph{pattern} of effects wrong (e.g., findings where humans show a large effect but the agent shows a small one, and vice versa), and an agent that gets the pattern right but produces effects on the wrong \emph{scale} (e.g., uniformly inflated tenfold). A plain Pearson correlation only catches the first---it is invariant to linear rescaling, so it would reward a tenfold-inflated agent as perfect. Lin's Concordance Correlation Coefficient~\citep{lawrence1989concordance} is the natural fix: it multiplies a Pearson-style pattern term by a bias-correction factor that explicitly penalizes mismatch in mean and variance. For a finding with $M$ tests, we collect human and agent effect-size vectors $\boldsymbol{\delta}_h, \boldsymbol{\delta}_a$ and compute:
\begin{equation}
\text{ECS}_{\text{finding}} = \rho \cdot C_b = \rho \cdot \frac{2\sigma_a\sigma_h}{\sigma_a^2 + \sigma_h^2 + (\mu_a - \mu_h)^2}
\end{equation}
where $\mu$ and $\sigma^2$ are the mean and variance of each effect-size vector. Here $\rho$ is the Pearson correlation, measuring precision (how well the agent captures the \textit{pattern} of effects across tests), while $C_b$ is the bias correction factor, penalizing systematic deviations in location and scale. $\text{ECS} \approx 1.0$ requires the agent to replicate both the relative structure and the exact magnitude of human effects.

\paragraph{Aggregation.} Because studies differ in the number of findings $N_s$, we set the weight of each finding $i$ from study $s$ to $w_i = 1/N_s$, so every study contributes equally. The global ECS is then computed as a study-balanced concordance over all finding-level effect sizes:
\begin{equation}
\text{ECS}_{\text{global}} = \frac{2 \sum w_i u_{a,i} u_{h,i}}{\sum w_i u_{a,i}^2 + \sum w_i u_{h,i}^2 + (\bar{\delta}_a - \bar{\delta}_h)^2}
\end{equation}
where $u_{a,i} = \delta_{a,i} - \bar{\delta}_a$ and $u_{h,i} = \delta_{h,i} - \bar{\delta}_h$ are centered deviations.

\subsection{Features of \textsc{HumanStudy-Bench}}
\label{subsec:features}
We summarize the key features of \textsc{HumanStudy-Bench} below.\\
\textbf{Optimization of agent design for participant simulation.} \textsc{HumanStudy-Bench} treats participant simulation in human studies as an \emph{agent design} problem: for each base model, it searches for the best agent design, revealing both models’ best-case performance and gaps to human behavior.\\
\textbf{High-fidelity human experiment reconstruction engine.}
\textsc{HumanStudy-Bench} reconstructs full human-subject experiments from the original papers and instantiates their protocols in a reusable engine, enabling faithful replication.\\
\textbf{Inferential-level alignment metrics.}
\textsc{HumanStudy-Bench} compares human and agent behavior at the level of inferential conclusions, applying the same analysis pipeline to both and summarizing heterogeneous statistical evidence into comparable alignment scores.\\
\textbf{Open platform for the research community.} \textsc{HumanStudy-Bench} is designed as an open platform that grows with community contribution. Practitioners can contribute validated published studies to benchmark AI agents, or design new experiments to formulate and pilot test their research hypotheses, evaluate agent designs before deployment, and systematically compare different agent specifications across diverse experimental paradigms.

\section{Experiment Setup}

We systematically disentangle the effects of base model capabilities, agent design specifications, and inference parameters on simulation fidelity.

\textbf{Human Studies.}
We evaluate against 12 human-subject studies (Appendix~\ref{app:study}) spanning individual cognition, strategic interaction, and social psychology. The individual cognition studies focus on cognitive biases and heuristic judgment~\citep{Ross1977TheC,Jacowitz1995MeasuresOA,Tversky1981TheFO,Kahneman1972SubjectivePA}. Strategic interaction studies are drawn from paradigms in game theory~\citep{Selten2007UnravelingIG,Shafir1992ThinkingTU,Forsythe1994FairnessIS,Berg1995TrustRA}. Social psychology studies examine social cognition, social norms, and group behavior~\citep{article,asch1946forming,gagnon1996discrimination,Prentice1993PluralisticIA}.

\textbf{Models and Inference Settings.}
We evaluate 10 contemporary models, including open-weight (e.g., Mistral, DeepSeek, Qwen) and proprietary APIs (e.g., Claude, GPT, Gemini, Grok). Model details in Appendix~\ref{app:model-id}. Motivated by work on diverse-model ensembles (e.g., MoE, multi-agent), we introduce a Mixed-Model Baseline: for each trial, we randomly sample one of 10 models to generate a response, repeating this 100 times. Unless otherwise noted in ablations, all models use a temperature $T=1.0$ to induce the behavioral variance needed for population simulation. For every study, the agent sample size is set to the \emph{original human sample size} reported in the source paper, which ranges from tens to over $2{,}100$ participants. Sample size governs statistical certainty---a smaller $N$ yields weaker evidence and a larger $N$ stronger evidence---so tying it to the original design preserves the experiment's power and avoids introducing a free parameter that could be tuned to manufacture or suppress significance. Given the large simulation sample sizes, the standard errors (SEs) are negligible ($\approx5\%$), so we omit them from the main results for clarity and report full SEs in Appendix~\ref{app:se}.

\textbf{Agent Design Variants.}
(1) \textit{Blank (A1)} uses the base model with no additional specification.
(2) \textit{Role-Play (A2)} instructs the model to act as a human participant in a psychological study, without specific attributes.
(3) \textit{Demographic (A3)} assigns attributes (e.g., age, gender, occupation) sampled from the original study's participant distribution.
(4) \textit{Contextualized Backstory (A4)} augments demographics with a rich natural language narrative about the agent's life history, personality, and daily context.
All four specifications share \emph{identical} task instructions and identical stimuli; the only component that varies is the identity block prepended to the system prompt (none, role only, role plus demographics, role plus demographics plus backstory). Differences across A1--A4 therefore isolate the effect of participant specification rather than of task framing. To confirm that the scores are not artifacts of instruction wording---for instance, prompt-induced demand characteristics that cue an agent toward the ``expected'' answer---we paraphrased all three prompt components while holding their informational content fixed: PAS moves by at most $8.8$ percentage points, against cross-specification differences exceeding $32$ points (Appendix~\ref{app:prompt-sensitivity}).
See Appendix~\ref{app:version} for the full templates.

\textbf{Evaluation Protocol.}
We deploy agents in reconstructed protocols to generate trial-level data, analyzed with the original statistical pipelines. We report the Probability Alignment Score (PAS; $[0,1]$) for inferential agreement and the Effect Consistency Score (ECS; $[-1,1]$) for effect-size alignment ($1$ indicates a perfect match to human data).

\section{Results}
\label{sec:results}

\begin{table}[t]\small
\centering
\caption{\text{Main Leaderboard.} Probabilistic Alignment (PAS) and Effect Consistency (ECS). Domain columns (Cog./Stra./Soc.) report PAS scores broken down by study domain. Best performing models are highlighted in \textcolor{best1}{\textbf{teal}}, worst in \textcolor{worst1}{\textbf{salmon}}. Cost breakdowns see Appendix~\ref{app:Moreexp}.}
\label{tab:pas-ecs-raw}
\setlength{\tabcolsep}{2pt}
\renewcommand{\arraystretch}{0.85}
\begin{tabular}{@{}
    >{\raggedright\arraybackslash}p{1.2cm}
    >{\centering\arraybackslash}p{0.7cm}
    >{\centering\arraybackslash}p{0.7cm}
    >{\centering\arraybackslash}p{0.95cm}
    >{\centering\arraybackslash}p{0.65cm}
    >{\centering\arraybackslash}p{0.85cm}
    >{\centering\arraybackslash}p{0.85cm}
    @{\hspace{6pt}\vrule\hspace{6pt}}
    >{\raggedright\arraybackslash}p{1cm}
    >{\centering\arraybackslash}p{0.7cm}
    >{\centering\arraybackslash}p{0.7cm}
    >{\centering\arraybackslash}p{0.95cm}
    >{\centering\arraybackslash}p{0.65cm}
    >{\centering\arraybackslash}p{0.85cm}
    >{\centering\arraybackslash}p{0.85cm}
@{}}
\toprule
\textbf{Model} & \textbf{} & \textbf{Cog.} & \textbf{Stra.} & \textbf{Soc.} & \textbf{PAS} & \textbf{ECS} &
\textbf{Model} & \textbf{} & \textbf{Cog.} & \textbf{Stra.} & \textbf{Soc.} & \textbf{PAS} & \textbf{ECS} \\
\midrule
\multirow{4}{*}{\shortstack[l]{Claude\\Haiku\\4.5}}
 & A1 & $0.35$ & \cellcolor{worst1}{$0.27$} & $0.47$ & $0.3616$ & $0.107$ &
\multirow{4}{*}{\shortstack[l]{GPT 5\\Nano}}
 & A1 & $0.33$ & $0.42$ & \cellcolor{worst3}{$0.33$} & $0.3594$ & \cellcolor{best2}{$0.253$} \\
 & A2 & $0.33$ & $0.29$ & $0.47$ & $0.3627$ & \cellcolor{best1}{$0.265$} &
 & A2 & $0.48$ & $0.31$ & \cellcolor{worst1}{$0.30$} & $0.3597$ & $0.220$ \\
 & A3 & $0.38$ & $0.32$ & $0.39$ & $0.3650$ & $0.202$ &
 & A3 & $0.45$ & $0.37$ & $0.45$ & $0.4226$ & \cellcolor{best4}{$0.243$} \\
 & A4 & $0.32$ & $0.29$ & \cellcolor{best2}{$0.65$} & $0.4181$ & $0.121$ &
 & A4 & \cellcolor{best2}{$0.64$} & $0.37$ & \cellcolor{worst2}{$0.33$} & $0.4442$ & $0.198$ \\
\midrule
\multirow{4}{*}{\shortstack[l]{DeepSeek\\V3.2}}
 & A1 & $0.35$ & $0.29$ & $0.38$ & \cellcolor{worst3}{$0.3405$} & $0.094$ &
\multirow{4}{*}{\shortstack[l]{GPT\\OSS\\120b}}
 & A1 & $0.26$ & $0.32$ & $0.46$ & \cellcolor{worst4}{$0.3471$} & $0.146$ \\
 & A2 & $0.49$ & \cellcolor{worst4}{$0.28$} & \cellcolor{worst4}{$0.34$} & $0.3701$ & $0.193$ &
 & A2 & $0.39$ & $0.30$ & $0.49$ & $0.3925$ & \cellcolor{worst4}{$0.064$} \\
 & A3 & $0.30$ & $0.31$ & $0.47$ & $0.3593$ & $0.148$ &
 & A3 & $0.39$ & $0.30$ & $0.47$ & $0.3889$ & $0.193$ \\
 & A4 & $0.36$ & $0.29$ & $0.52$ & $0.3859$ & $0.145$ &
 & A4 & $0.40$ & \cellcolor{worst3}{$0.28$} & $0.50$ & $0.3929$ & $0.132$ \\
\midrule
\multirow{4}{*}{\shortstack[l]{Gemini 3\\Flash}}
 & A1 & $0.36$ & $0.42$ & $0.46$ & $0.4155$ & $0.098$ &
\multirow{4}{*}{\shortstack[l]{GPT\\OSS\\20b}}
 & A1 & $0.59$ & $0.29$ & $0.51$ & $0.4641$ & $0.195$ \\
 & A2 & $0.32$ & $0.43$ & $0.45$ & $0.4005$ & $0.120$ &
 & A2 & $0.39$ & $0.31$ & $0.44$ & $0.3794$ & $0.171$ \\
 & A3 & $0.49$ & \cellcolor{best4}{$0.43$} & \cellcolor{best1}{$0.67$} & \cellcolor{best1}{$0.5314$} & $0.143$ &
 & A3 & $0.47$ & $0.31$ & $0.46$ & $0.4163$ & $0.218$ \\
 & A4 & $0.51$ & $0.35$ & $0.57$ & \cellcolor{best3}{$0.4771$} & $0.182$ &
 & A4 & $0.50$ & $0.29$ & $0.44$ & $0.4109$ & $0.212$ \\
\midrule
\multirow{4}{*}{\shortstack[l]{Mistral\\Nemo}}
 & A1 & $0.52$ & $0.32$ & $0.56$ & \cellcolor{best4}{$0.4663$} & $0.171$ &
\multirow{4}{*}{\shortstack[l]{Qwen 3\\Next80b}}
 & A1 & $0.31$ & $0.43$ & $0.44$ & $0.3914$ & $0.133$ \\
 & A2 & \cellcolor{best1}{$0.67$} & $0.32$ & $0.40$ & $0.4600$ & $0.179$ &
 & A2 & $0.24$ & $0.43$ & $0.43$ & $0.3694$ & $0.175$ \\
 & A3 & \cellcolor{best4}{$0.59$} & $0.32$ & $0.54$ & \cellcolor{best2}{$0.4829$} & $0.231$ &
 & A3 & $0.26$ & \cellcolor{best1}{$0.49$} & $0.45$ & $0.4012$ & $0.140$ \\
 & A4 & \cellcolor{best3}{$0.61$} & $0.28$ & $0.47$ & $0.4527$ & $0.219$ &
 & A4 & $0.36$ & $0.40$ & \cellcolor{best3}{$0.61$} & $0.4573$ & $0.122$ \\
\midrule
\multirow{4}{*}{\shortstack[l]{Gemma 4\\26b}}
 & A1 & \cellcolor{worst1}{$0.11$} & \cellcolor{best2}{$0.48$} & $0.46$ & $0.3516$ & \cellcolor{worst1}{$0.023$} &
\multirow{4}{*}{\shortstack[l]{Grok 4.1\\Fast}}
 & A1 & \cellcolor{worst3}{$0.19$} & $0.32$ & $0.46$ & \cellcolor{worst2}{$0.3214$} & $0.097$ \\
 & A2 & \cellcolor{worst4}{$0.23$} & \cellcolor{best3}{$0.46$} & $0.36$ & $0.3493$ & \cellcolor{worst3}{$0.063$} &
 & A2 & \cellcolor{worst2}{$0.12$} & $0.31$ & $0.44$ & \cellcolor{worst1}{$0.2891$} & $0.142$ \\
 & A3 & $0.28$ & $0.42$ & $0.42$ & $0.3752$ & $0.153$ &
 & A3 & $0.33$ & $0.31$ & \cellcolor{best4}{$0.58$} & $0.4074$ & \cellcolor{worst2}{$0.024$} \\
 & A4 & $0.31$ & $0.31$ & $0.54$ & $0.3874$ & \cellcolor{best3}{$0.245$} &
 & A4 & $0.38$ & \cellcolor{worst2}{$0.27$} & $0.41$ & $0.3524$ & $0.130$ \\
\midrule\midrule
\multirow{2}{*}{\shortstack[l]{Mixed\\Models}}
 & A1 & $0.01$ & $0.36$ & $0.43$ & $0.2706$ & $0.298$ &
\multirow{2}{*}{\shortstack[l]{Mixed\\Models}}
 & A3 & $0.20$ & $0.39$ & $0.40$ & $0.3293$ & $0.272$ \\
 & A2 & $0.19$ & $0.40$ & $0.41$ & $0.3331$ & $0.287$ &
 & A4 & $0.28$ & $0.39$ & $0.46$ & $0.3740$ & $0.247$ \\

\bottomrule
\end{tabular}
\end{table}

We evaluate the fidelity of LLM agents organizing 
results around three questions: how well agents reach the 
same inferential conclusions as humans (RQ1), how well 
their effect sizes match (RQ2), and how agent design 
choices shape the alignment (RQ3). Comprehensive
results see Appendix~\ref{app:hypothesis_details}.

\textbf{RQ1: Inferential alignment.}\\\textit{Can LLMs replicate the inferential conclusions of social science?}

\begin{figure}[h]
\centering
\includegraphics[width=\linewidth]{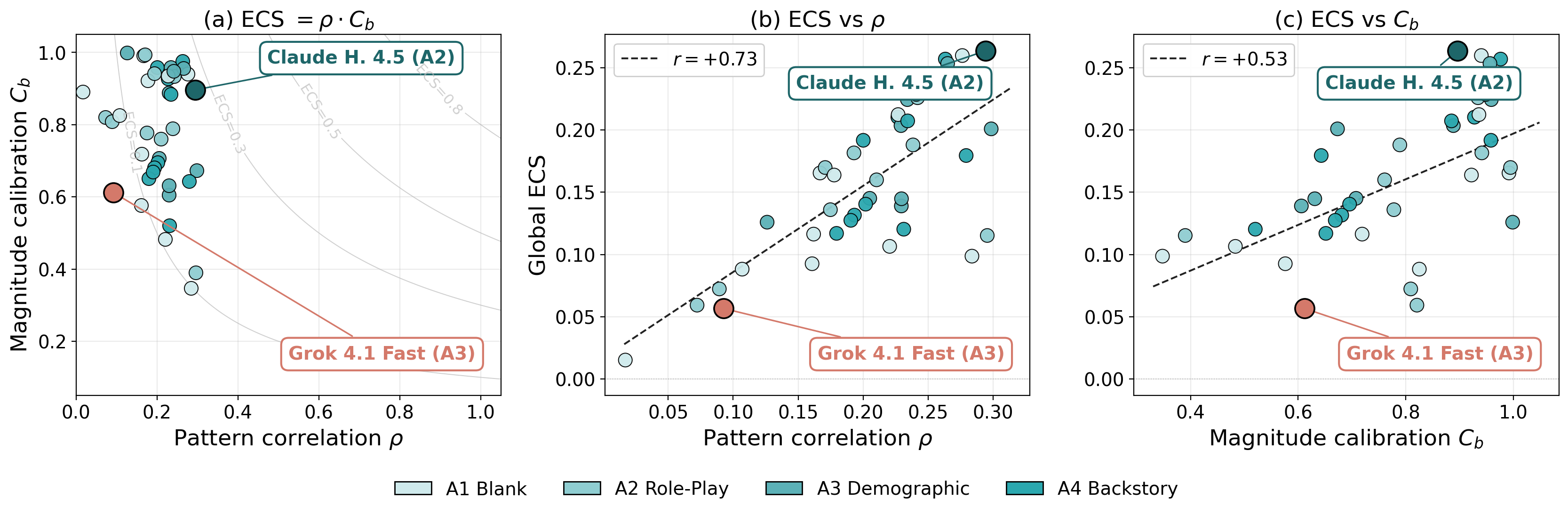}
\caption{\textbf{ECS decomposition into pattern correlation $\rho$ and magnitude calibration $C_b$.}
\textbf{(a)}~Each point is one agent plotted in the $(\rho, C_b)$ plane; gray contours are ECS isolines. Agents cluster in $\rho \le 0.31$, $C_b \in [0.35, 1.00]$. The leaderboard best (\textcolor[HTML]{1e6669}{Claude Haiku 4.5 A2}) and worst (\textcolor[HTML]{D4796A}{Grok 4.1 Fast A3}) are highlighted.
\textbf{(b, c)}~ECS plotted against $\rho$ and $C_b$ across all agents. ECS aligns more tightly with $\rho$ ($r=+0.73$) than with $C_b$ ($r=+0.53$), indicating that gains in pattern correlation translate more directly into ECS gains than improvements in magnitude calibration.}

\label{fig:ecs-decomposition}
\end{figure}

In Table~\ref{tab:pas-ecs-raw}, overall replication performance remains unsatisfactory across all evaluated models. Beyond this general limitation, the two metrics \emph{diverge}: the agent that best recovers human inferential conclusions is not the agent that best recovers human effect magnitudes. Gemini 3 Flash (A3) attains the highest alignment in the suite ({PAS $0.5314$}) at only moderate consistency ({ECS $0.143$}), indicating a model that reaches the right conclusion about whether an effect exists while systematically exaggerating or dampening its size. GPT 5 Nano (A1) shows the opposite profile: the second-highest consistency in the suite ({ECS $0.253$}) at one of the lowest alignments ({PAS $0.3594$}), indicating effect magnitudes that track the human pattern without accumulating enough evidence to reproduce the human significance conclusion. Neither profile dominates the other, and the two orderings are largely unrelated---agents are not arranged along a single axis of replication quality. We trace this divergence to the $\rho \cdot C_b$ decomposition of ECS below (Figure~\ref{fig:ecs-decomposition}): inferential alignment and magnitude alignment load on different components, so improving one does not automatically improve the other.

\begin{figure*}[h]
    \centering
    \includegraphics[width=\linewidth, trim={0 0 0 0}, clip]{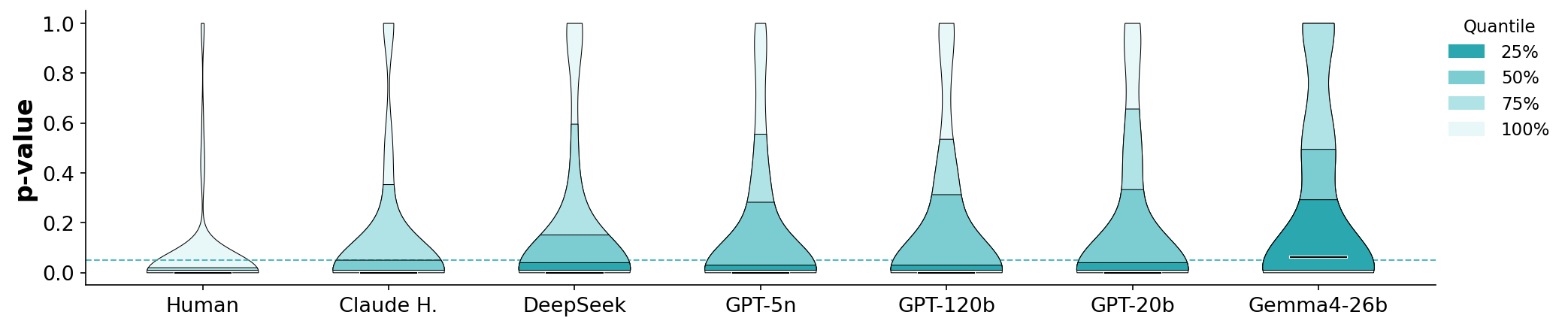}

    \caption{\textbf{Distribution of $p$-values across human and agents (A4 context).} The blue dashed line marks the standard statistical significance threshold ($p = 0.05$). The human data (far left) consistently yields highly significant results, with almost all probability mass tightly clustered near $p \approx 0$. In contrast, agent simulations fail to consistently reproduce these effects, showing a much wider spread of $p$-values with a substantial portion falling into the non-significant range ($p > 0.05$).}
    \label{fig:p_value_dist}
\end{figure*}

To understand the mechanics of these scores, we examine the distributional properties of the best-performing models. Figure~\ref{fig:p_value_dist} visualizes the $p$-value distributions underlying these scores. While human studies show a sharp peak at significance ($p < 0.05$), agent runs are dispersed, with substantial mass leaking into the non-significant range. To understand the mechanism behind this weak $p$-value pattern, we examine the PAS distribution at the hypothesis test level (Figure~\ref{fig:pas-violin}, Appendix~\ref{app:distributional_analysis}). We find a bimodal pattern---agents either closely replicate a human effect or miss it almost entirely---rather than a uniformly mediocre performance across hypothesis tests.

\begin{figure}[t]
\centering
\addtocounter{figure}{1}
\begin{minipage}[t]{0.48\linewidth}
  \centering
  \refstepcounter{subfigure}\label{fig:distributional-analysis}%
    (a) Effect-size correlation.\\[2pt]
  \includegraphics[width=\linewidth, trim={0 1em 0 0}, clip]{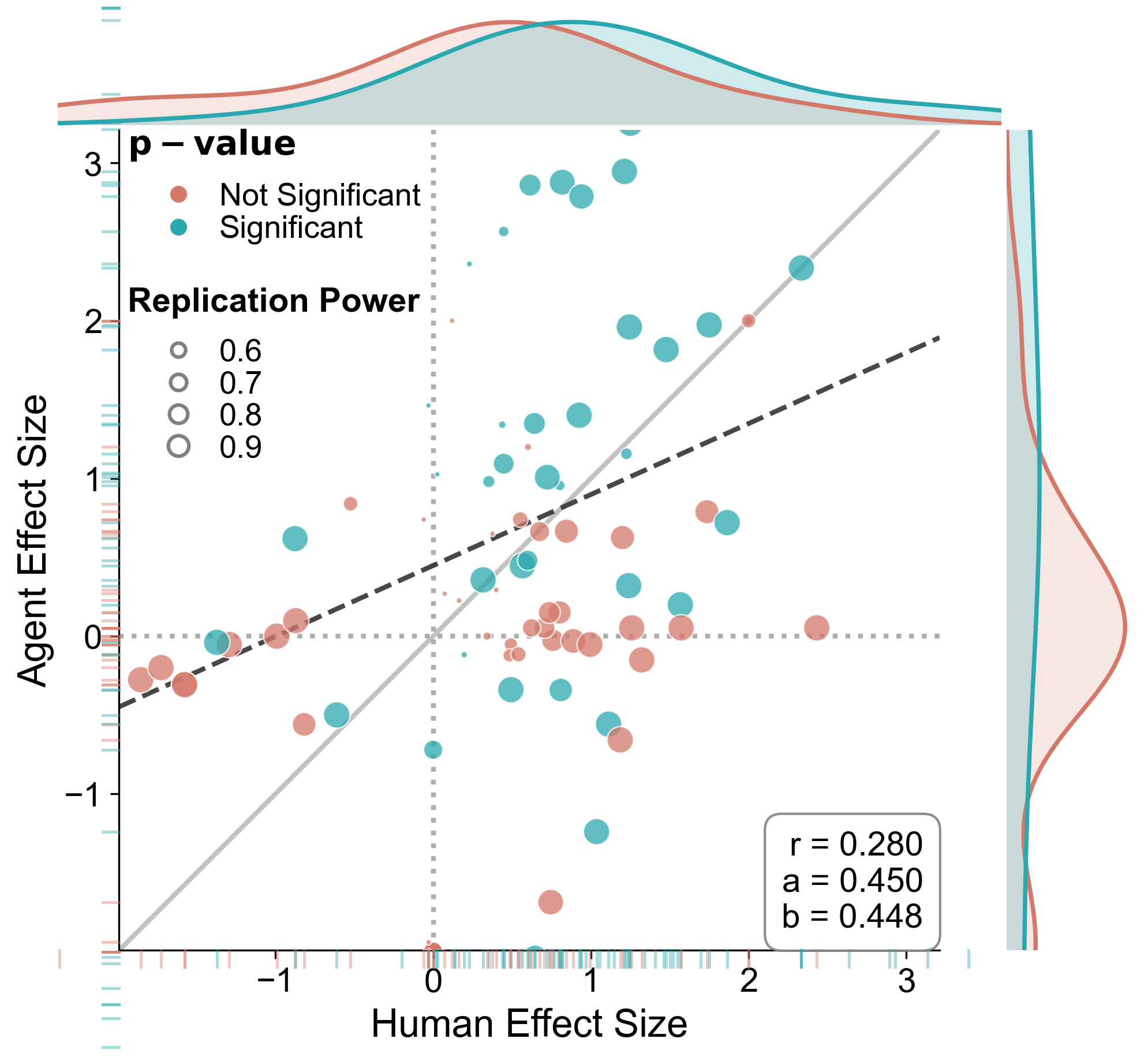}
\end{minipage}\hfill
\begin{minipage}[t]{0.48\linewidth}
  \centering
  \refstepcounter{subfigure}\label{fig:pas-violin}%
   (b) Test-level PAS distributions.\\[2pt]
  \includegraphics[width=\linewidth]{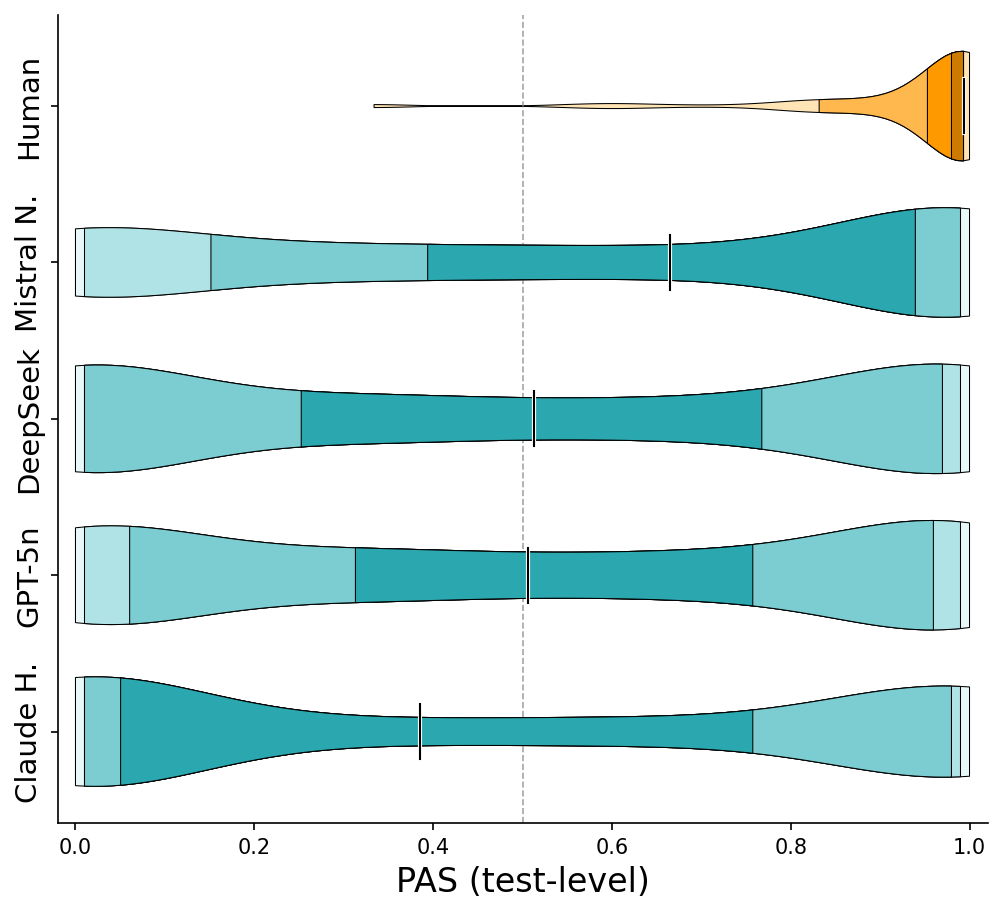}
\end{minipage}
\addtocounter{figure}{-1}
\caption{\textbf{Distributional analysis of simulation fidelity.} \textbf{(a)}~Effect-size correspondence for Claude Haiku 4.5 A2 vs.\ human. The gray diagonal marks perfect replication ($y{=}x$); the dashed line is the linear regression. Points are colored by significance (\textcolor[HTML]{2BA8AF}{significant}, $p<0.05$; \textcolor[HTML]{D4796A}{not significant}) and sized by replication power. Marginal densities show agent effect sizes are flatter and wider than human ones. \textbf{(b)}~Test-level PAS distributions. The human distribution (orange) is a baseline reference, showing that the selected tasks are robust in human populations. Agent distributions are bimodal---peaking near $0$ and $1$ with little intermediate mass---indicating agents either closely replicate a human effect or miss it almost entirely.}
\label{fig:per-test-alignment}
\end{figure}

\textbf{RQ2: Effect-size alignment and where it breaks.}\\\textit{Do agent effect-size magnitudes match human ones, and which part of the metric is the bottleneck?}

ECS scores in Table~\ref{tab:pas-ecs-raw} are uniformly low---the best agent reaches only $0.265$ (Claude Haiku 4.5, A2). Because ECS is bounded below $1$ by sampling noise in the human ground truth itself, this number is only interpretable against a ceiling. We therefore estimate what an idealized human replicator---one that matches the population effect exactly---would score under the same noise structure, by drawing paired replicates from each published effect's sampling distribution at the original sample size: this yields $\overline{\text{ECS}}_{\text{human}} \approx 0.928$ ($\rho \approx 0.932$, $C_b \approx 0.996$; Appendix~\ref{app:distributional_analysis}). The best agent thus attains roughly $29\%$ of the human ceiling, so the shortfall is a genuine gap rather than an unattainable target. The corresponding PAS ceiling is close to $1$: our filter admits only hypotheses with strong, independently replicated effects, so human evidence concentrates near $\pi_h \approx 1$ (orange reference in Figure~\ref{fig:pas-violin}), and the ambiguous case in which PAS hands an agent a free $0.5$ is rare by construction. Figure~\ref{fig:distributional-analysis} examines effect-size alignment in detail for the leaderboard best agent. In human-to-human replications, significant effects cluster along the diagonal~\cite{open2015estimating}; agent simulations, by contrast, show a more chaotic pattern: despite an overall dampening trend (regression slope $a = 0.526 < 1$), the low correlation ($r = 0.313$) indicates poor precision. The marginal distributions further show that human effect sizes are concentrated, whereas agent effect sizes are flatter and wider, confirming that agents tend to produce more extreme effects.

To localize the gap, we use the natural decomposition $\text{ECS} = \rho \cdot C_b$: $\rho$ is the cross-finding pattern correlation between agent and human effect sizes---intuitively, does the agent get the \emph{pattern} right (avoiding findings where humans show a large effect but the agent shows a small one, and vice versa)? $C_b$ is the magnitude calibration factor---intuitively, does the agent get the \emph{overall scale} right (effects centered, say, around $d \approx 1$ rather than blown up to $d \approx 10$)? A perfect replicator scores $\rho = C_b = 1$.

Figure~\ref{fig:ecs-decomposition}(a) plots every agent on the $(\rho,C_b)$ plane: every agent has $\rho \le 0.31$, while $C_b$ ranges up to $1.00$. The leaderboard best, Claude Haiku 4.5 (A2), already reaches a near-saturated $C_b = 0.90$ yet only $\rho = 0.29$; across all agents, ECS correlates more tightly with $\rho$ ($r = +0.73$, Figure~\ref{fig:ecs-decomposition}(b)) than with $C_b$ ($r = +0.53$, Figure~\ref{fig:ecs-decomposition}(c)). \textit{Pattern, not scale, is the bottleneck:} agents have already learned roughly the right magnitude of human effect sizes, but cannot reliably tell which findings should be larger than which. Concretely, $\rho$ asks whether the effects that are strong in humans are also the effects that are strong in agents; a value of at most $0.31$ means they largely are not. An agent can replicate several effects individually and still fail to order them---and that ordering is what a practitioner relies on when using simulated participants to decide which of several manipulations to pursue. This is precisely the failure mode PAS cannot see and ECS is designed to expose.

Two checks confirm that the low $\rho$ is not a measurement artifact. Recomputing ECS separately within each test family ($t$, $\chi^2$, $F$, correlation), with no cross-family conversion at all, leaves the values in a narrow $0.11$--$0.16$ band; and dropping each study in turn and recomputing leaves the global ECS within $0.160$--$0.195$, so no single study drives the score (Appendix~\ref{app:effect_heterogeneity}).

\textbf{RQ3: Does agent design close the gap?}\\\textit{How do design choices---context richness, temperature, and ensemble---affect simulation fidelity?}

\textit{Context helps, but not monotonically.} Adding demographic priors (A3) improves PAS over the blank baseline (A1) for $9/10$ models. Because the paired $t$-test over 10 models sits close to the threshold ($p = 0.049$), we report convergent evidence rather than resting on a single test: the Wilcoxon signed-rank test agrees ($p = 0.037$), and $9$ of $10$ models move in the same direction, so the conclusion depends neither on a distributional assumption nor on an individual influential model. Layering a richer backstory on top (A4) does not extend the trend: the global A4-vs-A3 difference fails to reach significance (paired $t$-test, $p = 0.82$), and per-model results bifurcate---A4 lifts models like Qwen3 Next 80b ($\Delta\text{PAS}=+0.056$) and Claude Haiku 4.5 ($+0.053$) but drags down Grok 4.1 Fast ($-0.055$) and Gemini 3 Flash ($-0.054$) (per-model contrasts in Appendix~\ref{app:hypothesis_details}). This suggests richer context 
helps up to a point, with a threshold that is model-specific rather 
than universal.

\begin{wraptable}{r}{0.55\textwidth}\small
\centering
\vspace{-1.2em}
\caption{\textbf{Temperature Ablation (Gemma 4 26b).} PAS / ECS across temperatures. Best in \textcolor{best1}{\textbf{teal}}, worst in \textcolor{worst1}{\textbf{salmon}}. Full per-domain results in Appendix~\ref{app:temp-ablation-full}.}
\label{tab:temp-ablation}
\setlength{\tabcolsep}{3pt}%
\renewcommand{\arraystretch}{0.85}%
\begin{tabular}{@{}l cccc | cccc@{}}
\toprule
 & \multicolumn{4}{c|}{\textbf{PAS}} & \multicolumn{4}{c}{\textbf{ECS}} \\
$T$ & A1 & A2 & A3 & A4 & A1 & A2 & A3 & A4 \\
\midrule
0.1 & \cellcolor{worst2}{.319} & .324 & .427 & \cellcolor{best3}{.471} & \cellcolor{best3}{.226} & .100 & .184 & .132 \\
0.3 & .336 & .381 & .395 & \cellcolor{best1}{.506} & .183 & \cellcolor{best1}{.251} & .189 & \cellcolor{worst3}{.068} \\
0.5 & .346 & .362 & .390 & \cellcolor{best2}{.488} & \cellcolor{best4}{.214} & .102 & .204 & \cellcolor{worst1}{-.009} \\
0.7 & \cellcolor{worst1}{.311} & .330 & \cellcolor{worst4}{.323} & \cellcolor{best4}{.431} & .147 & .188 & .172 & .072 \\
1.0 & \cellcolor{worst3}{.321} & .353 & .395 & .426 & \cellcolor{worst2}{.033} & \cellcolor{worst4}{.071} & .153 & \cellcolor{best2}{.245} \\
\bottomrule
\end{tabular}
\vspace{-1em}
\end{wraptable}

\textit{Temperature has no systematic effect on alignment.} Table~\ref{tab:temp-ablation} shows no monotone effect of temperature on PAS: within A4 alone, PAS swings non-monotonically across $T \in \{0.1, 0.3, 0.5, 0.7, 1.0\}$ ($0.43$--$0.51$), and the best temperature shifts across specifications ($T{=}0.3$ for A4, $T{=}0.5$ for A1). Decoding variance therefore behaves more like noise than a tunable knob; the structural gap to human behavior is not something we can sample our way out of (full domain-decomposed table in Appendix~\ref{app:temp-ablation-full}).

\textit{Naive model mixing does not improve alignment.} The Mixed-Models baseline (PAS $0.27$--$0.37$) is significantly below the single-model median (H4 in Appendix~\ref{app:hypothesis_details}, $p=0.009$). Mixing distinct response distributions adds within-condition noise without adding between-condition signal.

\textit{Model scale does not predict better alignment.} Flagship models do not dominate smaller open-weight ones 
(Table~\ref{tab:pas-ecs-raw}), and the largest swings 
come from changing the specification rather than the base 
model, suggesting that agent specification contributes 
more to alignment than model scale.



\textbf{Discussion.}
\textit{Memorization.} Pre-training exposure to these classic paradigms does not threaten our conclusions. Memorizing a conclusion is fundamentally different from reproducing a behavioral distribution across hundreds of independent trials, and memorization would inflate scores---yet mean single-model PAS is $0.397$, responses polarize bimodally, and the best-known strategic paradigms are the worst replicated ($0.343$ vs.\ $0.464$ for social psychology). Paraphrased and anonymized task variants move PAS by at most $5.2$ pp, and a model's ability to identify a study is dissociated from its ability to simulate it (Appendix~\ref{app:contamination}).

\textit{Why agents fail.} Since models overwhelmingly recognize these studies yet still fail to reproduce them, the failure is generative rather than a knowledge gap: agent effect sizes are flatter, wider, and more extreme than human ones (Figure~\ref{fig:distributional-analysis}), consistent with agents producing the responses of the participant they expect rather than the heterogeneous distribution a human population produces.

\textbf{Limitations.}
\label{limitation}
Our initial 12 hypotheses come from classic experiments (1946--2007) conducted on WEIRD (Western, educated, industrialized, rich, democratic) samples~\cite{Ross1977TheC,Jacowitz1995MeasuresOA,Tversky1981TheFO,Kahneman1972SubjectivePA,article,asch1946forming,Prentice1993PluralisticIA,Selten2007UnravelingIG,Shafir1992ThinkingTU,Forsythe1994FairnessIS,Berg1995TrustRA,gagnon1996discrimination}. Researchers using the initial 12 hypotheses to test human-likeness should interpret the scores with this scope in mind, particularly when evaluating agents meant to represent non-WEIRD or contemporary cohorts. We expect the contributions from more recent cross-cultural studies to broaden this coverage over time (Appendix~\ref{sec:impact}).
\section{Conclusion}
\label{sec:conclusion}

We propose validated behavioral hypotheses as a principled 
lens for evaluating human-likeness in LLM-based agents, and 
operationalize it through \textsc{HumanStudy-Bench}---an 
open platform that reconstructs published human-subject 
experiments end-to-end and administers the test to 
configurable agents. Our results reveal that current LLM 
agents fall short of human-likeness in specific, diagnostic 
ways. 
We hope \textsc{HumanStudy-Bench} can serve as a community infrastructure for objective, decomposable, and scalable evaluation of AI social simulations.



\bibliographystyle{tmlr}
\bibliography{neurips_2026}

\newpage

\appendix
\onecolumn
\section*{Contents}
\begin{description}
\item[A. Metrics Theoretical Foundation] \dotfill \pageref{app:metrics}
\begin{itemize}
    \item[A.1] Metric Intuition: What Are We Measuring? \dotfill \pageref{app:intuition}
    \item[A.2] Conditional Independence \dotfill \pageref{app:conditional}
    \item[A.3] Foundation of Probability Alignment Score \dotfill \pageref{app:theory}
\end{itemize}
\item[B. Methodological Rationale and Aggregation] \dotfill \pageref{app:methodology}
\begin{itemize}
    \item[B.1] Methodological Rationale \dotfill \pageref{app:rationale}
    \item[B.2] Metric Selection Rationale \dotfill \pageref{app:metric_selection}
    \item[B.3] Aggregation Implementation \dotfill \pageref{app:aggregation_impl}
    \item[B.4] Hierarchical Global Validity \dotfill \pageref{app:hierarchical_stats}
\end{itemize}
\item[C. Metrics Implementation Details] \dotfill \pageref{app:implementation}
\begin{itemize}
    \item[C.1] Evidence Transformation (Priors) \dotfill \pageref{app:prior-choice}
    \item[C.2] Generalization to Multiple Hypotheses \dotfill \pageref{app:generalization}
    \item[C.3] Standardized Effect Size Recovery \dotfill \pageref{app:effect_size}
\end{itemize}
\item[D. Implementation Details for Agent Design Variants] \dotfill \pageref{app:version}
\begin{itemize}
    \item[D.1] Blank (A1) \dotfill \pageref{subsec:v1}
    \item[D.2] Role-Play (A2) \dotfill \pageref{subsec:v2}
    \item[D.3] Demographic (A3) \dotfill \pageref{subsec:v3}
    \item[D.4] Contextualized Backstory (A4) \dotfill \pageref{subsec:v4}
\end{itemize}
\item[E. Summary of Studies] \dotfill \pageref{app:study}
\item[F. Complementary Experimental Results] \dotfill \pageref{app:Moreexp}
\begin{itemize}
    \item[F.1] Model names and OpenRouter identifiers \dotfill \pageref{app:model-id}
    \item[F.2] Bootstrap Standard Errors: Methodology and Justification \dotfill \pageref{app:se}
    \item[F.3] Extended Distributional Analysis \dotfill \pageref{app:distributional_analysis}
    \item[F.4] Hypothesis Testing Details \dotfill \pageref{app:hypothesis_details}
    \item[F.5] Contamination and Paradigm-Recognition Controls \dotfill \pageref{app:contamination}
    \item[F.6] Full Temperature Ablation \dotfill \pageref{app:temp-ablation-full}
    \item[F.7] Prompt Sensitivity Ablation \dotfill \pageref{app:prompt-sensitivity}
\end{itemize}  
\item[G. Broader Impacts] \dotfill \pageref{sec:impact}
\end{description}
\newpage
\section{Metrics Theoretical Foundation}
\label{app:metrics}
\subsection{Metric Intuition: What Are We Measuring?}
\label{app:intuition}

A replication study produces two kinds of evidence: a \textit{significance decision} (does the effect exist?) and an \textit{effect size} (how large is it?). These two quantities are related but not interchangeable: a large effect can fail to reach significance when the sample is small, and a tiny effect can be highly significant when the sample is massive. Our two metrics map directly onto this distinction.

\textbf{Probability Alignment Score (PAS) -- ``The Scientific Replication Rate.''}
PAS measures the \textbf{probability that the Agent and Humans agree on the scientific conclusion}---specifically, whether both sides find the effect significant or both find it non-significant. It answers the question: \textit{``If I use this Agent to replicate a hypothesis test, will I reach the same significance decision as the human study?''} PAS operates at the inferential level: it cares about the significance decision, not the exact numbers that produced it.

\textbf{Effect Consistency Score (ECS) -- ``The Data Fidelity.''}
ECS measures the \textbf{concordance of effect sizes} between Agent and Human data. It answers a complementary question: \textit{``Does the Agent reproduce the same pattern and magnitude of effects as Humans?''} ECS operates at the magnitude level: an agent that correctly identifies a ``framing effect'' but exaggerates it tenfold would score well on PAS but poorly on ECS.

The two metrics reflect complementary aspects of replication fidelity and are most informative when read together. High PAS with low ECS reveals an agent that reaches correct conclusions but with distorted magnitudes (e.g., extreme responses). Low PAS with moderate ECS reveals an agent whose effect sizes partially track human patterns but whose evidence falls short of statistical significance. Neither metric alone captures the full picture; their joint profile characterizes the agent's simulation capability.

\subsection{Conditional Independence}
\label{app:conditional}
The PAS derivation in Appendix~\ref{app:theory} requires one structural assumption. Let $\mathcal{H}$ and $\mathcal{A}$ denote the Human and Agent generative processes. We define the simulation task as estimating a latent truth parameter $\theta \in \{0, 1\}$ (where $1$ denotes the presence of an effect/Alternative Hypothesis, and $0$ denotes the Null) given a specific experimental design $\mathcal{E}$ (comprising stimuli, instructions, and conditions).

The core assumption enabling our framework is the \textbf{Conditional Independence} of the observation processes:
\begin{equation}
    P(y_h, y_a \mid \theta, \mathcal{E}) = P(y_h \mid \theta, \mathcal{E}) P(y_a \mid \theta, \mathcal{E})
\end{equation}
where $y_h$ and $y_a$ are the observable response data.

\textbf{Justification:} This independence holds because the Agent is not trained on the specific responses of the control group in the study being replicated; it generates behavior based solely on the semantic description of $\mathcal{E}$. Thus, given the ground truth $\theta$, the sampling noise in humans is independent of the stochastic decoding noise in the Agent.

\subsection{Foundation of Probability Alignment Score}
\label{app:theory}

We provide two theoretical interpretations of PAS. We begin with the Bayesian view, which directly derives the PAS formula from first principles, and then offer a Frequentist view that characterizes its statistical properties.

\paragraph{Perspective I: The Bayesian View (Minimum Bayes Risk).}
In the Bayesian ontology, $\theta$ are latent random variables. We seek an estimator $\hat{\mathcal{A}}$ that minimizes the expected error given the data. This perspective shows that PAS is not an ad hoc formula but the \textit{unique optimal} estimator under standard assumptions.

\textbf{1. Priors via Maximum Entropy.}
To avoid introducing subjective bias, we select priors based on the \textbf{Principle of Indifference}~\cite{jaynes2003probability}. For a binary state $\theta$, the distribution maximizing Shannon Entropy $H(\theta)$ is the uniform distribution, $P(\theta=1) = P(\theta=0) = 0.5$. This establishes the uninformative prior necessary for objective benchmarking.

\textbf{2. Minimizing Bayes Risk.}
We define the loss function as the Squared Error Loss with respect to the true alignment $A^*$: $\mathcal{L}(\hat{\mathcal{A}}, A^*) = (\hat{\mathcal{A}} - A^*)^2$.
The optimal estimator that minimizes the Bayes Risk (Expected Posterior Loss) is the conditional expectation (MMSE estimator):
\begin{equation}
    \hat{\mathcal{A}}_{Bayes} = \arg\min_{\hat{\mathcal{A}}} E_{\theta|D} [(\hat{\mathcal{A}} - A^*)^2] = E[A^* \mid D_h, D_a]
\end{equation}

\textbf{3. Derivation.}
Given the conditional independence of Human and Agent generative processes (Appendix~\ref{app:conditional}):
\begin{equation}
\begin{aligned}
    \hat{\mathcal{A}}_{Bayes} &= P(\theta_h = \theta_a \mid D_h, D_a) \\
    &= P(\theta_h=1|D_h)P(\theta_a=1|D_a) + P(\theta_h=0|D_h)P(\theta_a=0|D_a) \\
    &= \pi_h \pi_a + (1-\pi_h)(1-\pi_a) =\hat{\mathcal{A}}_{PAS}
\end{aligned}
\end{equation}

This creates a closed loop: the PAS formula presented in the main text is exactly the Minimum Bayes Risk estimator under Maximum Entropy priors.

\paragraph{Perspective II: The Frequentist View (Variance Reduction).}
In the Frequentist ontology, the latent truth states $\theta_h, \theta_a \in \{0, 1\}$ are fixed unknown constants. We aim to estimate the alignment indicator $\mathcal{A}^* = \mathbb{I}(\theta_h = \theta_a)$ based on observed data $D_h, D_a$.

Let $L = \ln \Lambda$ denote the Log-Likelihood Ratio derived from the data. By the Central Limit Theorem, the sampling distribution of $L$ is asymptotically normal: $L \sim \mathcal{N}(\mu, \sigma^2)$, where $\sigma$ represents sampling noise.

\textbf{1. The MLE Estimator (Hard Threshold).}
The Maximum Likelihood Estimator (MLE) for the alignment relies on the indicator function $\mathbb{I}(\cdot)$:
\begin{equation}
    \hat{\mathcal{A}}_{MLE} = \mathbb{I}(L_h > 0)\mathbb{I}(L_a > 0) + \mathbb{I}(L_h \le 0)\mathbb{I}(L_a \le 0)
\end{equation}
This estimator is unbiased asymptotically but exhibits maximal variance at the decision boundary ($L \approx 0$). Since $\hat{\mathcal{A}}_{MLE}$ behaves as a Bernoulli variable near the boundary, a marginal perturbation in noise causes a discrete jump, resulting in high instability:
\begin{equation}
    Var(\hat{\mathcal{A}}_{MLE}) \big|_{L \approx 0} = 0.25
\end{equation}

\textbf{2. The PAS Estimator (Soft Threshold).}
PAS can be understood as a \textbf{Shrinkage Estimator} using the logistic sigmoid function $\sigma(x) = (1+e^{-x})^{-1}$:
\begin{equation}
    \hat{\mathcal{A}}_{PAS} = \sigma(L_h)\sigma(L_a) + (1-\sigma(L_h))(1-\sigma(L_a))
\end{equation}
PAS serves as a continuous relaxation of MLE. Note that $\lim_{k \to \infty} \sigma(kx) = \mathbb{I}(x > 0)$; thus, PAS approaches MLE as evidence strength approaches infinity.

\textbf{3. Variance Reduction via Delta Method.}
We prove PAS reduces variance using the Delta Method approximation $Var(f(X)) \approx [f'(\mu)]^2 \sigma^2$. The derivative of the sigmoid at the boundary is $\sigma'(0) = 0.25$.
Comparing the variance of the decision component:
\begin{equation}
    Var(\hat{\mathcal{A}}_{PAS}) \big|_{L \approx 0} \approx [\sigma'(0)]^2 \sigma^2 = 0.0625 \sigma^2
\end{equation}
\textbf{Conclusion:} Provided the sampling noise is not catastrophic ($\sigma^2 < 4$), $Var(\hat{\mathcal{A}}_{PAS}) < Var(\hat{\mathcal{A}}_{MLE})$. PAS acts as a regularizer that trades a small bias (shrinkage towards 0.5) for a significant reduction in variance, minimizing the overall Mean Squared Error (MSE) in finite-sample regimes.

\section{Methodological Rationale and Aggregation}
\label{app:methodology}

\subsection{Methodological Rationale}
\label{app:rationale}
Our aggregation strategy and metric formulation diverge from standard meta-analytic approaches. We explicitly contrast our choices with alternative methodologies below.

\paragraph{Benchmarking vs. Meta-Analysis.}
While our framework aggregates results across multiple studies, we intentionally employ unweighted averaging rather than the precision-weighted averaging (inverse-variance weighting) typical of meta-analysis. This decision is grounded in two primary distinctions:

\begin{itemize}
    \item \textbf{Task Independence vs. Parameter Estimation:} Meta-analysis assumes that different studies estimate a shared biological parameter (e.g., a ``true'' population effect size). In contrast, our goal is benchmarking: evaluating an agent's general capability across a diverse suite of distinct tasks. Weighting by inverse variance would allow a single study with high statistical power to dominate the aggregate score, obscuring the agent's failure on smaller but equally critical tasks.

    \item \textbf{Avoiding Simulation Artifacts:} In participant simulation, the sample size of the agent ($N_{agent}$) is a controllable hyperparameter. Precision weighting would introduce a perverse incentive where the benchmark score is driven by the computational budget (generating more samples to artificially reduce variance) rather than behavioral fidelity. Unweighted averaging ensures the metric reflects average task performance, decoupled from simulation volume.
\end{itemize}

\subsection{Metric Selection Rationale}
\label{app:metric_selection}
We explicitly prioritize PAS over alternative metrics (e.g., raw effect size distance $|\delta_h - \delta_a|$ or distributional distances like Wasserstein) for two methodological reasons grounded in benchmarking rather than generative modeling:
\begin{enumerate}
    \item \textbf{Inferential Signal vs. Noise:} Human data contains variance from unobserved confounders irrelevant to the hypothesis. Distributional metrics prioritize matching this nuisance noise. PAS isolates the \textit{inferential signal}---the strength of evidence for the hypothesis---rewarding agents that capture the causal mechanism even if they exhibit less variance than humans.
    \item \textbf{Scale Invariance:} Effect sizes are scale-dependent (e.g., Cohen's $d$ vs. $\eta^2$). PAS normalizes these into a uniform probability space $[0,1]$, allowing aggregation across heterogeneous study designs.
\end{enumerate}

\subsection{Aggregation Implementation}
\label{app:aggregation_impl}
Scores are aggregated across the hierarchy (Test $\to$ Finding $\to$ Study $\to$ Benchmark) using variance-stabilizing transformations to ensure statistical robustness:

\begin{enumerate}
    \item \textbf{Finding \& Study Level (Variance Stabilization):} We map probabilities to correlation space ($r_j = 2\mathcal{A}_j-1$) and apply the Fisher-z transformation~\citep{fisher1921probable} to normalize the variance. Finding-level scores are the average of test $z$-scores; study-level scores are the average of finding $z$-scores. Both are mapped back to the $[0,1]$ PAS scale via the inverse hyperbolic tangent:
    \begin{equation}
        \bar{r} = \tanh \left( \frac{1}{M} \sum_{j=1}^M \text{arctanh}(r_j) \right), \quad \text{PAS} = (\bar{r} + 1)/2
    \end{equation}

    \item \textbf{Benchmark Level (Arithmetic Mean):} We compute the unweighted arithmetic mean of study-level PAS. Given the heterogeneity of the studies---which span diverse cognitive and social domains---we treat each study as an independent unit of capability. This approach ensures equal representation across domains and prevents any single study with distinct statistical properties (e.g., large $N$) from dominating the global benchmark score.
\end{enumerate}
ECS is aggregated similarly using study-balanced weights as described in the main text.

\subsection{Comparison with Standard Hypothesis Testing}
\label{app:hierarchical_stats}
A natural alternative to PAS is standard frequentist hypothesis testing: test whether the agent's effect sizes are statistically indistinguishable from the human's. We implemented this approach using a 4-tier hierarchical aggregation (Test $\to$ Finding $\to$ Study $\to$ Benchmark) and found that it is unsuitable as a benchmark metric: all agents yield $p < 0.001$, meaning the test rejects every model. This occurs because even small systematic deviations become highly significant given the large aggregate sample sizes. We report the procedure below for completeness.

\textbf{Level 1: Test-Level Standardization.}
For each individual test $k$ within finding $j$ of study $s$, we compute the standardized difference:
\begin{equation}
    Z_{s,j,k} = \frac{\hat{\delta}_{\text{agent}} - \hat{\delta}_{\text{human}}}{\sqrt{\text{SE}_{\text{agent}}^2 + \text{SE}_{\text{human}}^2}} \sim \mathcal{N}(0, 1)
\end{equation}

\textbf{Level 2: Finding-Level Aggregation.}
Because we are interested in the \textit{magnitude} of the discrepancy rather than its direction, we aggregate tests within finding $j$ using the Chi-squared statistic:
\begin{equation}
    \chi^2_{s,j} = \sum_{k=1}^{K_{s,j}} Z_{s,j,k}^2, \quad \text{with } p_{s,j} = 1 - F_{\chi^2_{K_{s,j}}}(\chi^2_{s,j})
\end{equation}
where $K_{s,j}$ is the number of tests in that finding.

\textbf{Level 3: Study-Level Aggregation (Stouffer).}
A study $s$ contains $m_s$ findings. To ensure each finding contributes equally regardless of its internal test count, we map the $p$-values to a standard normal space:
\begin{equation}
    Z_{s,j}^* = \Phi^{-1}(1 - p_{s,j}) \implies Z_{\text{study}, s} = \frac{1}{\sqrt{m_s}} \sum_{j=1}^{m_s} Z_{s,j}^*
\end{equation}

\textbf{Level 4: Benchmark-Level Aggregation.}
The final global $p$-value is:
\begin{equation}
    Z_{\text{benchmark}} = \frac{1}{\sqrt{S}} \sum_{s=1}^{S} Z_{\text{study}, s}, \quad P_{\text{global}} = 1 - \Phi(Z_{\text{benchmark}})
\end{equation}

\textbf{Result and Interpretation.}
This framework tests the Global Null Hypothesis ($H_0^{global}$) that $\delta_{\text{agent}} = \delta_{\text{human}}$ across all findings. In our experiments, all agents yield $P_{\text{global}} < 0.001$, confirming that no current LLM is statistically indistinguishable from humans. While this result is scientifically informative, it renders the test unsuitable as a \textit{benchmark metric}: a metric that assigns the same verdict to every model cannot discriminate between agent designs. PAS, by contrast, provides a continuous score that preserves ranking information.

\section{Metrics Implementation Details}
\label{app:implementation}

\subsection{Evidence Transformation (Priors)}
\label{app:prior-choice}
To compute posterior probabilities $\pi$, we calculate Bayes Factors ($BF_{10}$) using priors tailored to the test type. For t-tests and ANOVA, we employ the JZS prior (Cauchy distribution on effect size) with default scales $r=\sqrt{2}/2$ and $r=0.5$, respectively~\cite{rouder2009bayesian,rouder2012default}. These account for the majority of the tests. For contingency tables, we utilize a BIC-style approximation ($BF_{10} \approx \exp((\chi^2 - \text{df}\ln n)/2)$), while binomial tests use an exact conjugate Beta-Binomial prior (Beta(1,1)).

\paragraph{Sensitivity Analysis.}
\label{app:sensitivity}
To ensure that the benchmark rankings are driven by agent capability rather than specific prior choices in the evidence transformation, we conducted a sensitivity analysis on the prior scale $r$ used in the JZS Bayes Factor computation. The default value $r=0.707$ assumes a medium effect size distribution. We re-evaluated all agent outputs varying $r$ from $0.5$ (small effects) to $1.0$ (large effects).

As shown in Table~\ref{tab:sensitivity}, while the absolute magnitude of the posterior probabilities shifts slightly with the prior width, the relative ranking of agents remains highly stable (Spearman's $\rho > 0.99$). This confirms that PAS provides a consistent measure of relative model fidelity that is robust to reasonable variations in hyperparameter specification.

\begin{table}[h]
\centering
\caption{\textbf{Sensitivity Analysis of Cauchy Prior Scale ($r$) on Agent Rankings.} The high correlation ($\rho$) across scales indicates that the benchmark rankings are robust to the choice of prior.}
\label{tab:sensitivity}
\small
\begin{tabular}{lcccc}
\toprule
\textbf{Prior Scale ($r$)} & \textbf{Spearman's $\rho$} & \textbf{Mean $\Delta$ PAS} & \textbf{Max $\Delta$ PAS} & \textbf{Status} \\
\midrule
0.500 & 0.9992 & 0.000661 & 0.001738 & Stable \\
0.600 & 0.9998 & 0.000313 & 0.000824 & Stable \\
\textbf{0.707} (Default) & \textbf{1.0000} & \textbf{0.0000} & \textbf{0.0000} & \textbf{Baseline} \\
0.800 & 0.9998 & 0.000235 & 0.000657 & Stable \\
0.900 & 0.9997 & 0.000463 & 0.001307 & Stable \\
1.000 & 0.9996 & 0.000669 & 0.001906 & Stable \\
\bottomrule
\end{tabular}
\end{table}

\subsection{Generalization to Multiple Hypotheses}
\label{app:generalization}
Theoretically, PAS generalizes to $K$ hypotheses via the inner product of posterior vectors $\vec{\pi}_h \cdot \vec{\pi}_a$. In our implementation, we specifically operationalize this as a 3-way split ($H_+, H_-, H_0$). The resulting score is the dot product of the agent and human posterior vectors over these three outcome categories:
\begin{equation}
    S = \pi_{h+}\pi_{a+} + \pi_{h-}\pi_{a-} + \pi_{h0}\pi_{a0}
\end{equation}

\subsection{Standardized Effect Size Recovery}
\label{app:effect_size}
We recover Cohen's $d$ from reported statistics using the following conversions:
\begin{itemize}
    \item \textbf{T-family:} Independent t-tests use $d = t\sqrt{(n_1+n_2)/(n_1 n_2)}$; paired/one-sample tests use $d = t/\sqrt{n}$. F-tests ($df_1=1$) are converted to $t$-equivalents ($t=\sqrt{F}$) and processed similarly.
    \item \textbf{Correlation-family:} Pearson's $r$, Fisher's $z$, and Mann-Whitney $U$ (via rank-biserial $r_{rb}$) are converted to $d$ using the relationship $d = 2r/\sqrt{1-r^2}$.
    \item \textbf{Discrete:} $2 \times 2$ contingency tables are converted via the Log Odds Ratio ($d \approx \ln(OR)\sqrt{3}/\pi$). Binomial proportions use $d = 2(p - p_0)/\sqrt{p_0(1-p_0)}$.
\end{itemize}

\section{Implementation Details for Agent Design Variants}
\label{app:version}

We describe the implementation details for each agent design variant used in our experiments. Each variant corresponds to a different system prompt strategy that conditions the LLM's behavior during experimental participation. All four specifications share \emph{identical} task instructions---the two response-format lines reproduced in every template below---so the only component that varies across A1--A4 is the identity block (none, role only, role plus demographics, role plus demographics plus backstory). Differences in alignment across specifications therefore cannot be attributed to differences in task framing.

\subsection{Blank (A1)}
\label{subsec:v1}

The Blank variant serves as the baseline control condition. The model receives no identity block whatsoever---no role, no demographics, no backstory---allowing us to measure its intrinsic alignment with human behavior without any persona-based conditioning.

\begin{promptbox}[title={A1 System Prompt}]
\textit{(No identity block --- the model receives only the shared task instructions.)}

Follow the experimenter's instructions and answer each task in the requested format.\\
Be concise. Do not add extra explanations unless explicitly asked.
\end{promptbox}

\subsection{Role-Play (A2)}
\label{subsec:v2}

The Role-Play variant instructs the model to act as a human participant in a psychological study, but without assigning any specific demographic attributes. This test assesses whether the model has a generalizable concept of human experimental behavior.

\begin{promptbox}[title={A2 System Prompt}]
You are participating in a psychology experiment as a human participant.

Follow the experimenter's instructions and answer each task in the requested format.\\
Be concise. Do not add extra explanations unless explicitly asked.
\end{promptbox}

\subsection{Demographic (A3)}
\label{subsec:v3}

The Demographic variant augments the Role-Play prompt with specific demographic attributes (age, gender, education/background) sampled from the participant distribution reported in the original study. This tests whether models can condition their responses on population-level statistical priors.

\subsubsection{System Prompt Template}

\begin{promptbox}[title={A3 System Prompt Template}]
You are participating in a psychology experiment as a human participant.

YOUR IDENTITY:\\
- Age: \{age\} years old\\
- Gender: \{gender\}\\
- Education: \{education\}

Follow the experimenter's instructions and answer each task in the requested format.\\
Be concise. Do not add extra explanations unless explicitly asked.
\end{promptbox}

\subsubsection{Instantiated Example}

\begin{promptbox}[title={A3 Example (Instantiated)}]
You are participating in a psychology experiment as a human participant.

YOUR IDENTITY:\\
- Age: 21 years old\\
- Gender: Female\\
- Education: college student

Follow the experimenter's instructions and answer each task in the requested format.\\
Be concise. Do not add extra explanations unless explicitly asked.
\end{promptbox}

\subsection{Contextualized Backstory (A4)}
\label{subsec:v4}

The Contextualized Backstory variant extends the demographic profile with a rich, natural-language narrative describing the agent's life history, personality traits, relationships, and daily routines. This approach is inspired by the Generative Agents framework~\cite{park2023generativeagentsinteractivesimulacra}.

\subsubsection{Background Generation}

For each simulated participant, we generate a personalized backstory using an LLM (Gemini Flash). The generation prompt takes demographic information (name, age, gender, education, occupation) and produces a semicolon-delimited paragraph containing 5--6 statements about the agent's personality, routines, hobbies, living situation, and relationships.

\begin{promptbox}[title={Background Generation Prompt}]
Generate a life biography for \{name\}.

STYLE:\\
- Start with: `You are \{name\}.' then `\{name\} is ...'\\
- Single paragraph, semicolon-delimited statements\\
- 5-6 statements about: personality, routines, habits, hobbies, living situation, relationships\\
- NO experiments, studies, research, trials, or problem scenarios

DATA:\\
- Name: \{name\}, Age: \{age\}\\
- Gender: \{gender\}\\
- Education: \{education\}\\
- Occupation: \{occupation\}

EXAMPLE (John Lin):\\
John Lin is a pharmacy shopkeeper at the Willow Market and Pharmacy who loves to help people. He is always looking for ways to make the process of getting medication easier for his customers; John Lin is living with his wife, Mei Lin, who is a college professor, and son, Eddy Lin, who is a student studying music theory; John Lin loves his family very much; John Lin has known the old couple next-door, Sam Moore and Jennifer Moore, for a few years; John Lin thinks Sam Moore is a kind and nice man; John Lin knows his neighbor, Yuriko Yamamoto, well.

Generate bio for \{name\} (5-6 statements, pure life only, no experiments).
\end{promptbox}

\subsubsection{System Prompt Template}

The generated background is incorporated into the following system prompt template:

\begin{promptbox}[title={A4 System Prompt Template}]
You are participating in a psychology experiment as a human participant.

YOUR BACKGROUND AND MEMORIES:\\
\{generated\_background\}

Based on your background and memories above, respond as this participant would in the experiment.\\
Follow the experimenter's instructions and answer each task in the requested format.\\
Be concise. Do not add extra explanations unless explicitly asked.\\
Your responses should reflect your background, experiences, and characteristics as described above.
\end{promptbox}

\subsubsection{Instantiated Example}

\begin{promptbox}[title={A4 Example (Instantiated)}]
You are participating in a psychology experiment as a human participant.

YOUR BACKGROUND AND MEMORIES:\\
You are Christopher Hernandez. Christopher Hernandez is a dedicated landscape architect who finds peace in creating beautiful outdoor spaces for his local community; he starts every morning with a long walk through the neighborhood park to gather inspiration for his upcoming projects; Christopher Hernandez lives in a quiet suburban home with his wife, Elena, and their two teenage daughters who he treasures deeply; he spends most of his weekends woodworking in his garage or tending to his extensive backyard vegetable garden; Christopher Hernandez is known by his neighbors as a reliable and generous man who is always willing to lend a helping hand with home repairs; he maintains a close relationship with his younger brother, David, and they enjoy meeting up every Sunday for a round of golf.

Based on your background and memories above, respond as this participant would in the experiment.\\
Follow the experimenter's instructions and answer each task in the requested format.\\
Be concise. Do not add extra explanations unless explicitly asked.\\
Your responses should reflect your background, experiences, and characteristics as described above.
\end{promptbox}
\section{Summary of Studies}
\label{app:study}

\begin{table}[h]
\caption{Summary of studies used in this work.}
\label{tab:appendix_studies}
\centering
\small
\setlength{\tabcolsep}{3pt}
\renewcommand{\arraystretch}{1.15}
\begin{tabularx}{\linewidth}{p{0.09\linewidth} p{0.19\linewidth} X p{0.19\linewidth} p{0.07\linewidth}}
\toprule
\textbf{Category} & \textbf{Subdomain} & \textbf{Paper name} & \textbf{Author(s)} & \textbf{Year} \\
\midrule

\textbf{Individual Cognition} & False Consensus Effect & The ``False Consensus Effect'': An Egocentric Bias in Social Perception and Attribution Processes~\cite{Ross1977TheC} & Lee Ross; David Greene; Pamela House & 1977 \\
\addlinespace[0.6em]
 & Anchoring Effect & Measures of Anchoring in Estimation Tasks~\cite{Jacowitz1995MeasuresOA} & Karen E. Jacowitz; Daniel Kahneman & 1995 \\
\addlinespace[0.6em]
 & Framing Effect & The Framing of Decisions and the Psychology of Choice~\cite{Tversky1981TheFO} & Amos Tversky; Daniel Kahneman & 1981 \\
\addlinespace[0.6em]
 & Representativeness Heuristic & Subjective Probability: A Judgment of Representativeness~\cite{Kahneman1972SubjectivePA} & Daniel Kahneman; Amos Tversky & 1972 \\
\midrule

\textbf{Strategic Interaction} & p-Beauty Contest Game & Unraveling in Guessing Games: An Experimental Study~\cite{Selten2007UnravelingIG} & Rosemarie Nagel & 1995 \\
\addlinespace[0.6em]
 & Prisoner's Dilemma & Thinking through Uncertainty: Nonconsequential Reasoning and Choice~\cite{Shafir1992ThinkingTU} & Eldar Shafir; Amos Tversky & 1992 \\
\addlinespace[0.6em]
 & Ultimatum and Dictator Games & Fairness in Simple Bargaining Experiments~\cite{Forsythe1994FairnessIS} & Robert Forsythe; Joel L. Horowitz; N. E. Savin; Martin Sefton & 1994 \\
\addlinespace[0.6em]
 & Trust and Reciprocity Game & Trust, Reciprocity, and Social History~\cite{Berg1995TrustRA} & Joyce Berg & 1995 \\
\midrule

\textbf{Social Psychology} & Intentional Action and Side-Effects & Intentional Action and Side-Effects in Ordinary Language~\cite{article} & Joshua Knobe & 2003 \\
\addlinespace[0.6em]
 & Forming Impressions of Personality & Forming Impressions of Personality~\cite{asch1946forming} & S. E. Asch & 1946 \\
\addlinespace[0.6em]
 & Social Categorization & Discrimination in the Minimal Group Paradigm: Social Identity or Self-Interest?~\cite{gagnon1996discrimination} & Andr\'{e} Gagnon; Richard Y. Bourhis & 1996 \\
\addlinespace[0.6em]
 & Pluralistic Ignorance & Pluralistic Ignorance and Alcohol Use on Campus: Some Consequences of Misperceiving the Social Norm~\cite{Prentice1993PluralisticIA} & Deborah A. Prentice; Dale T. Miller & 1993 \\
\bottomrule
\end{tabularx}
\end{table}

\subsection{Study 1: False Consensus Effect}
\label{app:study1}
\textbf{Original study.}
Across three questionnaire substudies with Stanford undergraduates ($N=504$), participants made a choice and estimated what \% of peers would do the same. Substudy~1 used four hypothetical stories ($n=320$, $80$ per story) and included both peer-prevalence estimates and trait ratings of “typical” choosers; the false-consensus main effect was strong ($F(1,312)=49.1$) with parallel asymmetries in trait ratings ($F(1,312)=37.40$). Substudy~2 used a 35-item (34 analyzed) self-description checklist ($n=80$) where respondents self-categorized and estimated the \% of “college students in general” in their category. Substudy~3 used a hypothetical “sandwich-board” request with two sign versions ($n=104$ total) and again observed strong false consensus (combined $F=56.2$) alongside choice-consistent trait-rating differences (combined $F=17.79$)~\cite{Ross1977TheC}.

\textbf{Our reconstruction.}
We implement the same three substudies and prompt structure: four vignettes with choice + peer-prevalence estimates + trait ratings (Substudy~1), a 35-item (34 analyzed) self-categorization questionnaire with prevalence estimates (Substudy~2), and a two-version sandwich-board scenario with choice, prevalence estimates, and trait ratings (Substudy~3). We evaluate reconstructions by matching the paper’s reported aggregate targets (e.g., means and test statistics such as $F(1,312)=49.1$ and combined $F=56.2$), rather than original individual-level data, and we keep the participant pool as a generic “Stanford undergrad” (no added demographics).

\subsection{Study 2: Anchoring Effect}
\label{app:study2}
\textbf{Original study.}  
Jacowitz and Kahneman~\cite{Jacowitz1995MeasuresOA} quantified anchoring in numerical estimation using UC Berkeley students ($N=156$): a calibration group ($n=53$) first provided estimates and 10-point confidence ratings for 15 uncertain quantities, which defined low/high anchors at the 15th/85th percentiles. An anchored group ($n=103$) then judged whether each quantity was higher/lower than a provided anchor and gave an estimate plus confidence rating. Estimates shifted significantly toward anchors across all items.

\textbf{Our reconstruction.}  
We recreate the 15-item anchor--estimate procedure (higher/lower judgment $\rightarrow$ numeric estimate $\rightarrow$ 10-point confidence), and we implement both calibration-style baselines ($n\approx53$) and anchored conditions ($n\approx103$) using the same anchor percentiles. Evaluation targets the original aggregate patterns rather than individual-level replication.

\subsection{Study 3: Framing Effect }
\label{app:study3}
\textbf{Original study.}  
Tversky and Kahneman~\cite{Tversky1981TheFO} ran classroom questionnaire problems with students at Stanford University and the University of British Columbia. Sample sizes by problem were: Problem 1 (gain frame) $n=152$, Problem 2 (loss frame) $n=155$; Problem 3 $n=150$, Problem 4 $n=86$; Problem 5 $n=77$, Problem 6 $n=85$, Problem 7 $n=81$; Problem 8 $n=183$, Problem 9 $n=200$; Problem 10 version 1 $n=93$, version 2 $n=88$. 

\textbf{Our reconstruction.}  
We recreate the same set of decision vignettes (Problems 1--10) with identical choice structures and collect binary choices from participants. Evaluation focuses on reproducing the same directional reversals between matched frames (with the original problem-specific $n$ as targets), not exact replication of every reported percentage.

\subsection{Study 4: Representativeness Heuristic}
\label{app:study4}
\textbf{Original study.} 
This paper~\cite{Kahneman1972SubjectivePA} reported nine questionnaire substudies demonstrating representativeness-based judgments and sample-size neglect. Reported sample sizes were: Substudy 1 $n=92$, Substudy 2 $n=89$, Substudy 5 $n=52$, Substudy 6 $N=1500$, Substudy 7 $n=97$, Substudy 8 $N=560$, Substudy 9 $n=115$; Substudies 3 and 4 did not report $N$ in the extracted design summary.

\textbf{Our reconstruction.} 
We recreate all nine substudies with the original wording and response formats, and we evaluate against the paper’s aggregated outcomes using the same substudy sample sizes where available (Substudies 1,2,5,6,7,8,9). We do not generate new individual-level datasets; replication is scored by matching the reported summary judgments/medians and direction of errors.

\subsection{Study 5: p-Beauty Contest Game}
\label{app:study5}
\textbf{Original study.}  
Nagel~\cite{Selten2007UnravelingIG} investigated iterated reasoning and convergence in p-beauty contest guessing games. Participants repeatedly chose a number in the interval $[0,100]$, aiming to be closest to $p$ times the group mean, with $p = 1/2, 2/3,$ or $4/3$. Choices across four rounds were recorded to assess reasoning depth and adjustment over time. Results showed bounded rationality: initial choices clustered around a few iteration steps from 50, and repeated play led choices to move toward the Nash equilibrium.

\textbf{Our reconstruction.}  
We reimplement the p-beauty contest as a standardized multi-round numeric guessing task with the same $p$ conditions. Participants provide guesses and receive round-wise feedback in a uniform interface. Evaluation targets clustering and directional convergence rather than exact numerical distributions.

\subsection{Study 6: Prisoner’s Dilemma  }
\label{app:study6}
\textbf{Original study.}  
Shafir and Tversky~\cite{Shafir1992ThinkingTU} tested nonconsequential reasoning with Princeton undergraduates across three tasks: Prisoner’s Dilemma (PD) triads ($n=80$), a computerized Newcomb’s problem ($n=40$), and a PD information-seeking variant ($n$ not reported). In the PD triads, participants completed 40 one-shot games (6 PDs) presented in three versions (opponent unknown/known, compete/known, cooperate/known), totaling 444 PD triads. Outcomes showed higher cooperation when the opponent’s choice was unknown and a majority one-box preference in Newcomb’s problem.

\textbf{Our reconstruction.}  
We recreate the same three tasks as standardized vignettes: PD triads (444 triads; $n=80$), Newcomb’s problem ($n=40$), and the PD information-seeking variant. Participants make the same discrete choices in each scenario, and the evaluation targets the same directional patterns.

\subsection{Study 7: Ultimatum and Dictator Games}
\label{app:study7}
\textbf{Original study.}  
Forsythe et al.\cite{Forsythe1994FairnessIS} tested fairness in bargaining with University of Iowa students ($N=230$) across six between-subjects experiments: $5$ Dictator–Pay ($n=45$), $5$ Ultimatum–Pay ($n=43$), $5$ Dictator–NoPay ($n=46$), $5$ Ultimatum–NoPay ($n=48$), $10$ Dictator–Pay ($n=24$), $10$ Ultimatum–Pay ($n=24$). Proposers chose an allocation; in Ultimatum games responders could reject (both get $0$). Offers were higher in Ultimatum than Dictator, and Dictator offers were higher under NoPay than Pay. 

\textbf{Our reconstruction.}  
We implement the same six conditions ($n=45,43,46,48,24,24$) as standardized allocation/acceptance tasks (Dictator: allocate, Ultimatum: allocate + accept/reject), but without lab payments. Evaluation targets the same directional contrasts (Ultimatum $>$ Dictator; NoPay Dictator $>$ Pay Dictator).

\subsection{Study 8: Trust and Reciprocity Game}
\label{app:study8}
\textbf{Original study.}  
Berg et al.\cite{Berg1995TrustRA} studied trust and reciprocity in a two-stage investment game with University of Minnesota undergraduates ($N=120$): No-History ($n=64$, 32 pairs) and Social-History ($n=56$, 28 pairs). Room A chose how much of a \$10 endowment to send (\$0--\$10); the amount was tripled; Room B decided how much to return. Social history consisted of a report summarizing the prior 32 pairs’ outcomes.

\textbf{Our reconstruction.}  
We recreate the same two conditions (No-History $n=64$, Social-History $n=56$) with identical send/return rules and the same social-history report structure. Evaluation targets the original directional effects. 

\subsection{Study 9: Intentional Action and Side-
Effects}
\label{app:study9}
\textbf{Original study.}  
Knobe~\cite{article} tested the side-effect effect in two between-subjects vignette experiments with people in a Manhattan public park ($N=120$). Experiment~1 (chairman/environment) had $N=78$ ($n=39$ harm, $n=39$ help): 82\% judged the harmful side effect intentional vs.\ 23\% in the helpful condition. Experiment~2 (lieutenant/soldiers) had $N=42$ ($n=21$ harm, $n=21$ help): 77\% vs.\ 30\% intentional ($\chi^2(1,N=42)=9.5$). 

\textbf{Our reconstruction.}  
We recreate both vignettes with the same harm/help conditions and the same response formats: Yes/No intentionality plus 0--6 blame/praise ratings (Exp~1), and Yes/No intentionality (Exp~2). We evaluate by reproducing the harm--help asymmetry in intentionality judgments using the original sample sizes ($N=78$, $N=42$).

\subsection{Study 10: Forming Impressions of Personality}
\label{app:study10}
\textbf{Original study.}  
Asch~\cite{asch1946forming} reported impression-formation experiments with college students (total $N=811$) using short trait lists. Centrality manipulations were tested in Experiment~I ($N=166$: warm $n=90$, cold $n=76$) and Experiment~III ($N=46$: polite $n=20$, blunt $n=26$). Primacy manipulations were tested by reversing list order in Experiment~VI ($N=58$: order A $n=34$, order B $n=24$) and Experiment~VII ($N=99$: order A $n=46$, order B $n=53$). Impressions were measured via selected traits/ratings, showing strong central-trait and order effects.

\textbf{Our reconstruction.}  
We recreate Experiments~I, III, VI, and VII as standardized text vignettes and collect the same structured trait judgments, targeting the original between-condition sample sizes: Exp~I ($90$ and $76$), Exp~III ($20$ and $26$), Exp~VI ($34$ and $24$), Exp~VII ($46$ and $53$). Evaluation focuses on reproducing the centrality (warm/cold) and primacy (order) directional effects.

\subsection{Study 11: Social Categorization}
\label{app:study11}
\textbf{Original study.}
Gagnon and Bourhis~\cite{gagnon1996discrimination} tested minimal-group discrimination with undergraduates ($N=94$) in a 2$\times$2 design: Interdependence (autonomous vs.\ interdependent) $\times$ In-group Identification (low vs.\ high, measured post-hoc). Participants were randomly assigned to groups by coin toss and allocated resources using Tajfel matrices. Contrary to the behavioral interaction model, autonomous individuals discriminated as much as interdependent respondents (no main effect of Interdependence, $F<1.56$, $p>.05$). In line with social identity theory, high in-group identifiers discriminated significantly more than low identifiers (FAV on MJP: $F(1,90)=18.51$, $p<.001$).

\textbf{Our reconstruction.}
We recreate the same 2-factor design ($N=94$) with text-based coin-toss categorization (no visual stimuli), six Tajfel matrices (three types $\times$ original and reversed forms), a 100-point zero-sum distribution, and a post-allocation identification measure. The interdependence factor is manipulated via instructions; identification is measured after allocation and used for post-hoc median split, matching the original procedure.

\subsection{Study 12: Pluralistic Ignorance}
\label{app:study12}
\textbf{Original study.} 
Surveys with Princeton undergraduates tested pluralistic ignorance about campus drinking (total $N=468$ across the three studies we reconstructed). Study~1 ($n=132$) used 11-point comfort ratings for \emph{self} and the \emph{average student} (plus an IQR bracket for “50\% of students”). Study~2 ($n=242$) added \emph{friends} and manipulated question order (self-first vs.\ other-first) on the same 11-point scale. Study~4 ($n=94$) focused on the keg-ban policy (0--10 attitude scale) and measured perceived deviance plus social-action and alienation indicators. All showed the same pattern: students rated themselves as less comfortable (or less aligned with the norm) than they believed others were, and this misperception was associated with lower action and greater alienation~\cite{Prentice1993PluralisticIA}. 

\textbf{Our reconstruction.} 
We rebuild these three studies using the original items and response scales (Study~1 $n=132$, Study~2 $n=242$, Study~4 $n=94$), keeping the same target comparisons (self vs.\ average student; friends; order; keg-ban deviance/action/alienation). Analyses rely on the paper’s reported summary statistics; we do not simulate individual-level data or recreate the longitudinal Study~3.

\section{Complementary Experimental Results}
\label{app:Moreexp}

\subsection{Model names and OpenRouter identifiers}
\label{app:model-id}
All single-model evaluations use the OpenRouter API. Table~\ref{tab:model-openrouter} lists the display name used in this report and the exact OpenRouter model ID (provider/series) used for each run.

\begin{table*}[h]
\centering
\caption{Model display names and OpenRouter model IDs. All runs use the OpenRouter API; the ``OpenRouter model ID'' is the exact \texttt{model} string sent to the API.}
\label{tab:model-openrouter}
\begin{tabular}{@{}ll@{}}
\toprule
\textbf{Model} & \textbf{OpenRouter model ID} \\
\midrule
Claude Haiku 4.5   & \texttt{anthropic/claude-haiku-4.5} \\
DeepSeek V3.2      & \texttt{deepseek/deepseek-v3.2} \\
Gemini 3 Flash     & \texttt{google/gemini-3-flash-preview} \\
Mistral Nemo       & \texttt{mistralai/mistral-nemo} \\
Gemma 4 26b        & \texttt{google/gemma-4-26b-a4b-it} \\
GPT 5 Nano         & \texttt{openai/gpt-5-nano} \\
GPT OSS 120b       & \texttt{openai/gpt-oss-120b} \\
GPT OSS 20b        & \texttt{openai/gpt-oss-20b} \\
Qwen3 Next 80b     & \texttt{qwen/qwen3-next-80b-a3b-instruct} \\
Grok 4.1 Fast      & \texttt{x-ai/grok-4.1-fast} \\
\bottomrule
\end{tabular}
\end{table*}

\subsection{Bootstrap Standard Errors: Methodology and Justification}
\label{app:se}

Uncertainty for PAS (Probability of Alignment with Science) is quantified via a \emph{participant-level bootstrap} standard error (SE). Because we already have large sample sizes, re-testing each agent produces negligible final score deviation. The procedure is as follows.

\begin{table*}[h]\small
\centering
\caption{\textbf{Total PAS and Total SE.} Each cell reports Total PAS with its bootstrap Total SE in parentheses; Total PAS is the simple mean of per-study raw PAS, and Total SE is propagated from per-study bootstrap SEs as $\sqrt{\sum \mathrm{SE}_i^2}/N$.}
\label{tab:pas-pas-agg-horizontal}
\setlength{\tabcolsep}{4pt}
\renewcommand{\arraystretch}{1.0}
\begin{tabular}{@{}l cccc@{}}
\toprule
\textbf{Model} & \textbf{A1} & \textbf{A2} & \textbf{A3} & \textbf{A4} \\
\midrule
Claude Haiku 4.5 & $0.3616$ {\scriptsize(0.0057)} & $0.3627$ {\scriptsize(0.0088)} & $0.3650$ {\scriptsize(0.0081)} & $0.4181$ {\scriptsize(0.0076)} \\
DeepSeek V3.2 & $0.3405$ {\scriptsize(0.0082)} & $0.3701$ {\scriptsize(0.0090)} & $0.3593$ {\scriptsize(0.0067)} & $0.3859$ {\scriptsize(0.0066)} \\
Gemini 3 Flash & $0.4155$ {\scriptsize(0.0042)} & $0.4005$ {\scriptsize(0.0003)} & $0.5314$ {\scriptsize(0.0061)} & $0.4771$ {\scriptsize(0.0041)} \\
Mistral Nemo & $0.4663$ {\scriptsize(0.0102)} & $0.4600$ {\scriptsize(0.0126)} & $0.4829$ {\scriptsize(0.0121)} & $0.4527$ {\scriptsize(0.0121)} \\
Gemma 4 26b & $0.3516$ {\scriptsize(0.0028)} & $0.3493$ {\scriptsize(0.0074)} & $0.3752$ {\scriptsize(0.0087)} & $0.3874$ {\scriptsize(0.0079)} \\
GPT 5 Nano & $0.3594$ {\scriptsize(0.0114)} & $0.3597$ {\scriptsize(0.0102)} & $0.4226$ {\scriptsize(0.0107)} & $0.4442$ {\scriptsize(0.0071)} \\
GPT OSS 120b & $0.3471$ {\scriptsize(0.0139)} & $0.3925$ {\scriptsize(0.0104)} & $0.3889$ {\scriptsize(0.0123)} & $0.3929$ {\scriptsize(0.0079)} \\
GPT OSS 20b & $0.4641$ {\scriptsize(0.0129)} & $0.3794$ {\scriptsize(0.0123)} & $0.4163$ {\scriptsize(0.0126)} & $0.4109$ {\scriptsize(0.0105)} \\
Qwen3 Next 80b & $0.3914$ {\scriptsize(0.0039)} & $0.3694$ {\scriptsize(0.0057)} & $0.4012$ {\scriptsize(0.0088)} & $0.4573$ {\scriptsize(0.0070)} \\
Grok 4.1 Fast & $0.3214$ {\scriptsize(0.0045)} & $0.2891$ {\scriptsize(0.0039)} & $0.4074$ {\scriptsize(0.0049)} & $0.3524$ {\scriptsize(0.0088)} \\
\midrule
Mixed Models & $0.2706$ {\scriptsize(0.0146)} & $0.3331$ {\scriptsize(0.0133)} & $0.3293$ {\scriptsize(0.0129)} & $0.3740$ {\scriptsize(0.0088)} \\
\bottomrule
\end{tabular}
\end{table*}

\paragraph{Resampling.}
For each study and each model--method combination, the PAS score is a function of participant-level outcomes (e.g., test statistics and significance flags). We take the participant pool as the empirical distribution and draw $B$ bootstrap samples \emph{with replacement} of the same size as the original sample. The reported results use $B=200$ unless stated otherwise.

\paragraph{Bootstrap distribution and SE.}
On each bootstrap sample we recompute the study-level PAS (or the same metric used in the main analysis). The bootstrap distribution of that metric is summarized by its standard deviation, which we take as the \emph{bootstrap standard error} for that study. Formally, $\widehat{\mathrm{SE}}_{\mathrm{boot}} = \mathrm{SD}(\hat\theta^*_1,\ldots,\hat\theta^*_B)$, where $\hat\theta^*_b$ is the statistic from the $b$-th bootstrap sample. No distributional assumption is made beyond the data-generating process implied by the empirical participant set.

\paragraph{Aggregation across studies.}
For each model--method, a single ``Total PAS'' is computed as the mean of per-study PAS over the $K$ studies. Under the assumption that study-level estimates are approximately independent, the SE of the mean is propagated as
\[
  \widehat{\mathrm{SE}}(\overline{\mathrm{PAS}}) = \frac{1}{K}\sqrt{\sum_{k=1}^K \widehat{\mathrm{SE}}_k^2}\,.
\]

\subsection{Extended Distributional Analysis}
\label{app:distributional_analysis}

In the main text (RQ1), we identified a distinct bimodal signature in agent alignment: agents tend to either closely replicate a human effect or miss it almost entirely, with little intermediate mass (Figure~\ref{fig:pas-violin} in the main text). Here, we extend this analysis to determine whether this polarization stems from benchmark insensitivity (i.e., tasks being exclusively ``too easy'' or ``too hard'') or intrinsic model behaviors. Our auxiliary analysis confirms that the evaluation suite maintains high discriminatory power, effectively disentangling model capabilities.

\begin{figure}[t]
    \centering
    \includegraphics[width=\linewidth]{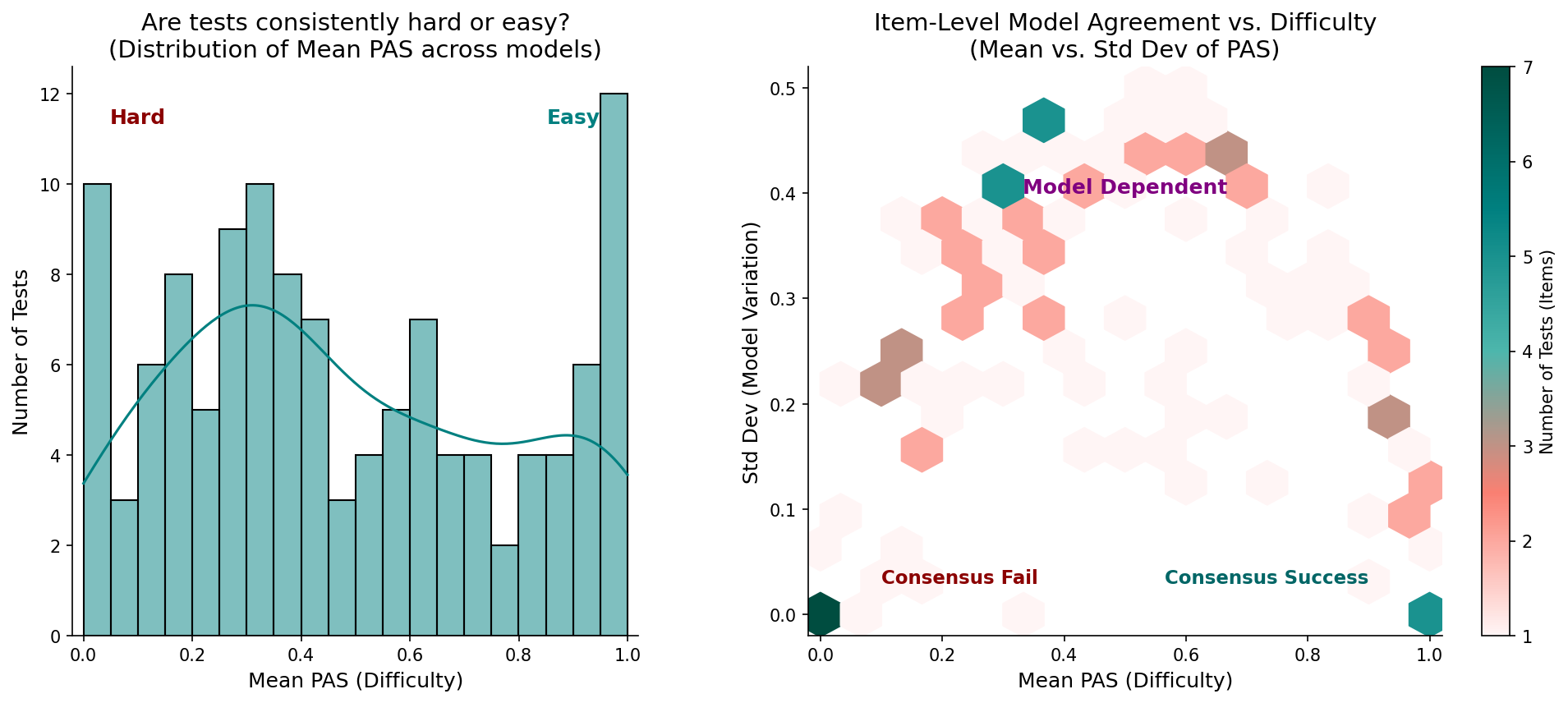} 
    \caption{\textbf{Decomposing Variance via Task Difficulty.} \textbf{(Left)} The benchmark spans a wide spectrum of difficulty levels, refuting the notion that tasks are binary. \textbf{(Right)} The Hexbin analysis identifies a ``Zone of Disagreement'' (red center), where variance peaks. This indicates that the benchmark effectively disentangles models: in this zone, architectural choices—rather than task difficulty alone—determine success or failure.}
    \label{fig:task_difficulty}
\end{figure}

\paragraph{Benchmark Sensitivity and Model Disentanglement}
To interrogate the source of variance, we decompose performance by task difficulty in Figure~\ref{fig:task_difficulty}. The distribution of mean PAS (Left) demonstrates that the benchmark covers a broad difficulty spectrum rather than collapsing into a binary distribution.

Crucially, the mean-variance analysis (Right) reveals a ``Zone of Disagreement'' at medium difficulty levels. The existence of this high-variance zone confirms that the benchmark effectively disentangles models based on their architectural priors. The bimodal outcome observed in the main text is therefore not an artifact of task homogeneity (with only hard and easy task for all), but a result of Agents being heterogeneous.

\begin{figure}[h]
    \centering
    \includegraphics[width=0.9\linewidth]{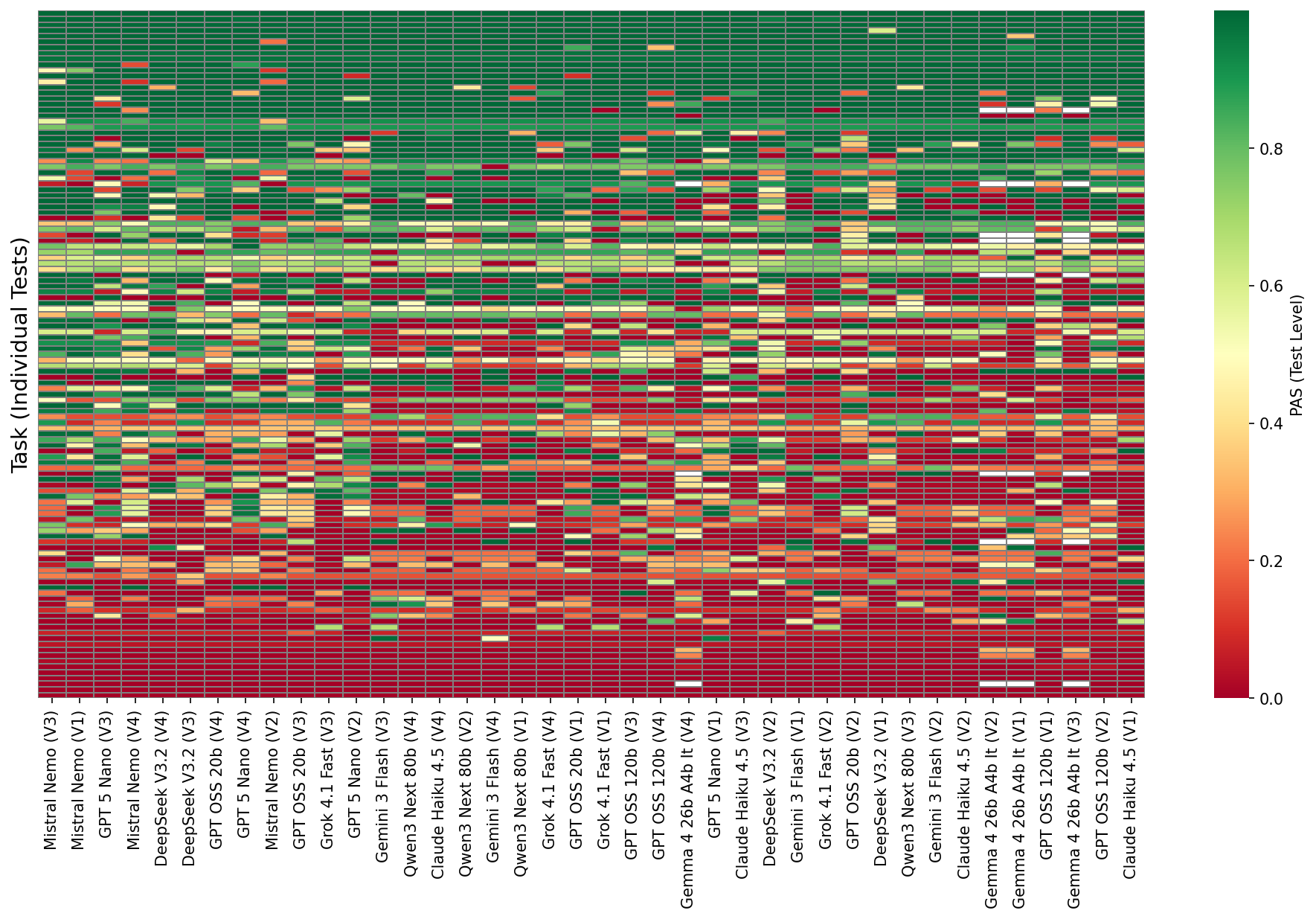}
    \caption{\textbf{The Landscape of Idiosyncratic Capabilities.} The scattered distribution of high (green) and low (red) alignment scores illustrates that capabilities are fragmented. No single agent universally dominates; instead, performance is highly specific to the interaction between agent and task type.}
    \label{fig:heatmap_matrix}
\end{figure}

\paragraph{Idiosyncratic Capabilities}
Finally, Figure~\ref{fig:heatmap_matrix} visualizes the item-level performance across all model-prompt combinations. The lack of continuous vertical bands (which would indicate uniform dominance by a single model) highlights the idiosyncratic nature of current simulation capabilities. We observe a ``patchwork'' pattern where all model configurations exhibit sporadic divergences on tasks that otherwise other models solve. This suggests that ``General Simulation Intelligence'' is not yet linear; rather, different architectures offer complementary strengths in modeling specific facets of human behavior.

\paragraph{Robustness to Effect-Size Heterogeneity.}
\label{app:effect_heterogeneity}
A natural concern is that the low cross-finding pattern correlation $\rho$ in our ECS decomposition (\S\ref{sec:results}, RQ2) could stem from converting heterogeneous effect-size metrics ($d$, $h$, $\eta^2$, $r$) onto a common scale (Appendix~\ref{app:effect_size}). To rule this out, we recompute ECS separately within each test family ($t$, $\chi^2$, $F$, correlation), avoiding any cross-family conversion. The resulting ECS values stay in a narrow $0.11$--$0.16$ band, matching the global ECS magnitudes in Table~\ref{tab:pas-ecs-raw}. A second concern operates at the study level rather than the metric level: a single idiosyncratic study could dominate a pooled concordance score. To rule this out, we drop each study in turn and recompute the global ECS; it stays within $0.160$--$0.195$, so no individual study drives the result. The low pattern correlation therefore reflects a genuine simulation gap, not a measurement artifact.

We note that pooling heterogeneous paradigms onto one standardized scale is the established construction in replication research rather than a choice specific to this work: the Reproducibility Project~\cite{open2015estimating} placed roughly 100 studies from very different paradigms on a common effect-size scale and summarized the field with a single correlation between original and replication effects. ECS applies the same construction to agent-versus-human effects, with two differences: we use Cohen's $d$ as the common scale, converting each test with standard meta-analytic formulas (Appendix~\ref{app:effect_size}), and we use Lin's concordance coefficient in place of Pearson, because concordance additionally requires the magnitudes to match. Normalizing the effect sizes---standardizing agent and human effects to zero mean and unit variance---would reduce the metric exactly to $\rho$, which is one of the two factors of ECS and is reported separately in Figure~\ref{fig:ecs-decomposition}.

\paragraph{Is $0.26$ Small? A Human-Side Noise Ceiling.}
A further concern is that ECS may be bounded well below $1$ by the sampling noise in the human ground truth itself, in which case the agent's best-case ECS of $0.265$ might already be close to the achievable ceiling rather than far from it. To quantify this, we construct a \emph{human-side upper bound}: for each of the $43$ findings with a recoverable Cohen's $d$, we treat the published effect size as a Gaussian centered on the reported point estimate with standard deviation given by the standard sampling-noise approximation $\mathrm{SE}(d) \approx \sqrt{4/n + d^2/(2n)}$ at the original human sample size $n$, draw two independent human-side replicates per finding, and compute ECS (Lin's CCC) between them. Averaging over $B = 1000$ such draws gives an estimate of the ECS that an idealized human replicator---one that exactly matches the population effect---would attain under the noise structure of our benchmark. This yields an upper bound of $\overline{\mathrm{ECS}}_{\text{human}} \approx 0.928$, with $\rho \approx 0.932$ and $C_b \approx 0.996$. The best agent's $\mathrm{ECS} = 0.265$ thus sits at roughly $29\%$ of the human noise ceiling: the gap to humans is genuine, not an artifact of an unattainable target.
\subsection{Hypothesis Testing Details}
\label{app:hypothesis_details}

\begin{table}
\caption{\textbf{Summary of Hypothesis Tests.} Results based on aggregated PAS scores. Significance levels: $^{*}p<0.05$, $^{**}p<0.01$, $^{***}p<0.001$.}
\label{tab:hypothesis_summary}
\centering
\small
\begin{tabular}{llcccc}
\toprule
\textbf{ID} & \textbf{Hypothesis Description} & \textbf{Method} & \textbf{$\Delta$} & \textbf{Statistic} & \textbf{Result} \\
\midrule
\multicolumn{6}{l}{\textit{Group A: Agent Design}} \\
H1 & A3 (Demo) $>$ A1 (Empty) & Paired $t$ & $+0.033$ & $t=2.27^{*}$ & \textbf{Supported} \\
H2 & A3 (Demo) $>$ A4 (Narrative) & Paired $t$ & $-0.003$ & $t=-0.23$ & Null \\
H3 & A1 $>$ A2 (Role-Play) & Paired $t$ & $+0.009$ & $t=0.77$ & Rejected \\
H4 & Median $>$ Mixed Model & Pooled $t$ & $+0.070$ & $t=2.73^{**}$ & \textbf{Supported} \\
\midrule
\multicolumn{6}{l}{\textit{Group B: Domain Specificity}} \\
H5 & Social $>$ Cognition & Pooled $t$ & $+0.080$ & $t=3.24^{**}$ & \textbf{Supported} \\
H6 & Social $>$ Strategic & Pooled $t$ & $+0.121$ & $t=7.25^{***}$ & \textbf{Supported} \\
\midrule
\multicolumn{6}{l}{\textit{Group C: Model Scaling}} \\
H7 & Flagship $\neq$ Open & Mann-Whitney & N/A & $p=0.91$ & Null \\
\midrule
\multicolumn{6}{l}{\textit{Group D: Hyperparameters (Ablation)}} \\
H8 \& H9 & PAS at $T=0.1$ vs $T=1.0$ & Wilcoxon & $+0.006$ & $p=1.00$ & Null \\
\bottomrule
\end{tabular}
\end{table}

We evaluate our ten core hypotheses using a combination of pooled $t$-tests (incorporating Standard Errors of PAS) and Wilcoxon signed-rank tests for robustness. Table~\ref{tab:hypothesis_summary} summarizes the statistical outcomes.

\paragraph{Group A: Agent Design (RQ2)}
Our analysis confirms that specific prompt engineering strategies significantly impact simulation fidelity, though not always monotonically.

\begin{itemize}
    \item \phantomsection\def\thecurrentHypothesis{H1}\label{hyp:H1} \textbf{H1: Demographic Benefit (Supported).} Adding demographic descriptors (A3) yields a statistically significant improvement in PAS compared to the blank baseline (A1) (paired $t$-test $p=0.049$; Wilcoxon $p=0.037$). This effect is robust, with 9 out of 10 models showing positive gains.

    \item \label{hyp:H2}\textbf{H2: Narrative Overload (Null).} We fail to reject the null hypothesis that complex backgrounds (A4) and simple demographics (A3) yield equivalent PAS (paired $t$-test $p=0.824$, $n=10$). There is no aggregate directional preference: 6 of 10 models prefer A4 and 4 prefer A3, with a mean PAS difference of essentially zero ($\Delta = -0.003$). Per-model variation is large---A4 lifts Qwen3 Next 80b ($\Delta\text{PAS}=+0.056$) and Claude Haiku 4.5 ($+0.053$) but drags down Grok 4.1 Fast ($-0.055$) and Gemini 3 Flash ($-0.054$)---consistent with the model-specific bifurcation noted in the main text.

    \item \label{hyp:H3} \textbf{H3: Role-Play Penalty (Rejected).} Globally, instructing models to ``act as a human'' (A2) does not significantly degrade PAS compared to implicit conditioning (A1) (paired $t$-test $p=0.46$). The earlier ``Rationality Bias'' pattern is confined to specific models (Claude Haiku, Grok); on the new study suite no global penalty survives.

    \item \label{hyp:H4}\textbf{H4: Consistency Necessity (Supported).} The Mixed-Models Baseline performs significantly worse than the median single-model approach (single mean $0.397$ vs.\ mixed mean $0.327$; pooled $t$-test $p=0.009$). This confirms that aggregating diverse response distributions introduces ``destructive interference,'' diluting the systematic behavioral signals required for valid replication.
\end{itemize}

\paragraph{Group B: Domain Robustness (RQ3)}
On the current study suite, models perform best on social-psychology tasks and worst on strategic (game-theory) tasks. Cognitive-bias tasks fall in between. Note that since PAS is not directly comparable among different fields, we use raw study-level PAS averaged within each domain.

\begin{itemize}
    \item \label{hyp:H5}\textbf{H5: Social vs.\ Cognition (Supported).} Models perform better on Social Psychology than on Cognitive Bias tasks (Soc PAS$=0.464$ vs.\ Cog PAS$=0.384$; $t$-test $p=0.002$).

    \item \label{hyp:H6}\textbf{H6: Social vs.\ Strategic (Strongly Supported).} Models perform substantially better on Social tasks than on Strategic ones (Soc PAS$=0.464$ vs.\ Stra PAS$=0.343$; $t$-test $p<10^{-9}$). Strategic interaction tasks (e.g., Trust Game, Beauty Contest) remain the hardest behavioral domain to replicate.
\end{itemize}

\paragraph{Group C \& D: Scaling and Hyperparameters (RQ3)}
We observe that raw capability metrics do not directly translate to behavioral simulation fidelity.

\begin{itemize}
    \item \label{hyp:H7}\textbf{H7: Model Scale (Null).} Contrary to established scaling laws, expensive ``Flagship'' models do not significantly outperform efficient or open-weights models in reproducing human behavior (Flagship best-PAS mean $0.450$ vs.\ Open best-PAS mean $0.428$; Mann-Whitney $p=0.91$).

    \item \label{hyp:H910}\textbf{H8 \& H9: Temperature Ablation (Null).} Variance scaling has a minimal effect on replication fidelity. Neither high temperature ($T=1.0$) nor low temperature ($T=0.1$) provided a statistically significant advantage in alignment or consistency (Wilcoxon $p=1.0$), suggesting the simulation gap is structural rather than stochastic.
\end{itemize}

\subsection{Contamination and Paradigm-Recognition Controls}
\label{app:contamination}

Every paradigm in the suite is canonical, so training exposure is plausible by default. This appendix reports the two controls summarized in the Discussion paragraph of Section~\ref{sec:results}: a task-rewriting control that asks whether performance survives when surface form is changed, and a recognition probe that asks whether a model's ability to identify a study predicts its ability to simulate it.

\paragraph{Paraphrase and anonymization.}
We constructed two variants of each study's materials. \emph{Paraphrasing} rewrites syntax and wording while preserving every entity, quantity, and response option. \emph{Anonymization} replaces named referents (people, places, institutions) with generic descriptions while preserving the domain and narrative structure. Both preserve the experimental logic, since altering it would forfeit the human ground truth that makes the comparison a replication. We held fixed the agent specification, participant profiles, sample size, temperature, base model, evaluator, and human reference data, varying only the materials. We evaluated 4 models on 2 studies under 3 conditions with 3 seeds each, for 72 runs in total; the two studies were chosen to span the range of baseline difficulty, one near the top of the suite and one near the bottom.

\begin{table}[h]\small
\centering
\caption{\textbf{PAS under task-rewriting controls.} Paraphrasing and anonymization change PAS by at most $5.2$ percentage points, on both an easy and a hard study. Values are means over 4 models $\times$ 3 seeds.}
\label{tab:contamination-controls}
\begin{tabular}{@{}lccc@{}}
\toprule
\textbf{Study} & \textbf{Original} & \textbf{Paraphrased} & \textbf{Anonymized} \\
\midrule
Side-effect effect~\cite{article} & $0.968$ & $0.924$ ($-4.4$ pp) & $0.916$ ($-5.2$ pp) \\
Trust Game~\cite{Berg1995TrustRA} & $0.352$ & $0.321$ ($-3.1$ pp) & $0.345$ ($-0.8$ pp) \\
\bottomrule
\end{tabular}
\end{table}

Performance is essentially unchanged under both controls (Table~\ref{tab:contamination-controls}). Notably, the effect is small in both directions of the difficulty range: the easy study does not collapse when its surface form is rewritten, and the hard study does not improve. If recognition of specific wording were driving replication, we would expect the rewritten variants to degrade substantially.

\paragraph{Recognition probe.}
The rewriting control leaves open a stronger version of the objection: that agents recognize the \emph{paradigm} rather than the wording. We therefore measured recognition directly. After administering the study materials, we asked each model whether it recognized the source and could name the original authors or paradigm, and scored the proportion of runs in which it did so correctly. If recognition drove replication, recognition rate and PAS would covary.

\begin{table}[h]\small
\centering
\caption{\textbf{Recognition is dissociated from simulation quality.} Recognition rate (\% of runs correctly identifying the source study) against PAS on the same study. Knowing the study is neither necessary (GPT-OSS-120b on Knobe) nor sufficient (both models on Berg) for reproducing its participants' behavior.}
\label{tab:recognition-probe}
\begin{tabular}{@{}lcccc@{}}
\toprule
& \multicolumn{2}{c}{\textbf{Knobe (2003)}} & \multicolumn{2}{c}{\textbf{Berg (1995)}} \\
\cmidrule(lr){2-3}\cmidrule(lr){4-5}
\textbf{Model} & \textbf{Recognition} & \textbf{PAS} & \textbf{Recognition} & \textbf{PAS} \\
\midrule
DeepSeek V3.2 & $100\%$ & $0.998$ & $95\%$ & $0.293$ \\
GPT OSS 120b  & $5\%$   & $0.997$ & $65\%$ & $0.353$ \\
\bottomrule
\end{tabular}
\end{table}

They do not covary (Table~\ref{tab:recognition-probe}). On Knobe's side-effect vignette, one model identifies the source in every run and the other in $5\%$ of runs, yet both replicate the effect almost perfectly---recognition is not necessary. On the Trust Game, both models identify the paradigm most of the time and both fail to replicate it---recognition is not sufficient. The two abilities are dissociated, which is what the memorization account has to explain.

\paragraph{Scope of these controls.}
These controls cover a subset of the suite rather than all 12 studies, and by construction they preserve task structure, quantities, and response options. They therefore do not exclude every channel through which prior exposure might shift a response distribution. What they do establish, together with the aggregate evidence reported in the main text---mean PAS $0.397$, $\rho \le 0.31$, and the best-known paradigms being the worst replicated---is that recognition does not account for the observed pattern of failures. Any residual benefit from exposure would raise agent scores, making the reported agent--human gap a conservative one.

\subsection{Full Temperature Ablation}
\label{app:temp-ablation-full}

Table~\ref{tab:temp-ablation-full} expands the main-text temperature ablation (Table~\ref{tab:temp-ablation}) by decomposing PAS into the three pre-registered domains---Cognition (Cog.), Strategic Interaction (Stra.), and Social Psychology (Soc.)---across all four agent specifications (A1--A4) and five temperatures ($T \in \{0.1, 0.3, 0.5, 0.7, 1.0\}$) for Gemma 4 26b. Two patterns are worth noting. First, the optimal temperature is \emph{domain-dependent}: cognition tasks favor low-to-moderate temperatures ($T \le 0.5$), strategic-interaction tasks peak at $T = 1.0$ for several specs (e.g., A1, A2, A3), and social-psychology tasks have no consistent optimum. This heterogeneity is masked when only the aggregate PAS is reported. Second, ECS varies non-monotonically with $T$ and is not aligned with PAS---e.g., A4 attains its best PAS at $T = 0.3$ but its best ECS at $T = 1.0$---reinforcing the divergence between the two metrics discussed in Section~\ref{sec:results}. Overall, no single temperature dominates across domains or specifications, indicating that temperature tuning trades off cognitive precision against strategic and social variance rather than producing a uniform improvement.

\begin{table*}[h]\small
\centering
\caption{\textbf{Full Temperature Ablation (Gemma 4 26b).} Per-domain and total PAS/ECS across temperatures. Within each specification, \textcolor{best1}{\textbf{teal}} marks the best temperature and \textcolor{worst1}{\textbf{salmon}} marks the worst (per column).}
\label{tab:temp-ablation-full}
\begin{tabular}{@{}ll cccccc@{}}
\toprule
\textbf{Spec} & \textbf{T} & \textbf{Cog.} & \textbf{Stra.} & \textbf{Soc.} & \textbf{PAS} & \textbf{ECS} \\
\midrule
\multirow{5}{*}{A1} & 0.1 & 0.23 & 0.32 & \cellcolor{best1}{0.41} & 0.3193 & \cellcolor{best1}{0.226} \\
 & 0.3 & 0.25 & 0.41 & \cellcolor{worst1}{0.35} & 0.3359 & 0.183 \\
 & 0.5 & 0.24 & 0.41 & 0.39 & \cellcolor{best1}{0.3456} & 0.214 \\
 & 0.7 & \cellcolor{best1}{0.35} & \cellcolor{worst1}{0.21} & 0.37 & \cellcolor{worst1}{0.3112} & 0.147 \\
 & 1.0 & \cellcolor{worst1}{0.11} & \cellcolor{best1}{0.45} & 0.40 & 0.3213 & \cellcolor{worst1}{0.033} \\
\midrule
\multirow{5}{*}{A2} & 0.1 & 0.24 & 0.32 & 0.41 & \cellcolor{worst1}{0.3243} & 0.100 \\
 & 0.3 & 0.34 & 0.32 & 0.48 & \cellcolor{best1}{0.3808} & \cellcolor{best1}{0.251} \\
 & 0.5 & 0.26 & 0.32 & \cellcolor{best1}{0.51} & 0.3622 & 0.102 \\
 & 0.7 & \cellcolor{best1}{0.37} & \cellcolor{worst1}{0.21} & 0.41 & 0.3304 & 0.188 \\
 & 1.0 & \cellcolor{worst1}{0.23} & \cellcolor{best1}{0.46} & \cellcolor{worst1}{0.37} & 0.3526 & \cellcolor{worst1}{0.071} \\
\midrule
\multirow{5}{*}{A3} & 0.1 & 0.39 & 0.35 & \cellcolor{best1}{0.54} & \cellcolor{best1}{0.4268} & 0.184 \\
 & 0.3 & 0.38 & 0.32 & 0.49 & 0.3954 & 0.189 \\
 & 0.5 & \cellcolor{best1}{0.42} & 0.34 & 0.42 & 0.3904 & \cellcolor{best1}{0.204} \\
 & 0.7 & 0.37 & \cellcolor{worst1}{0.20} & \cellcolor{worst1}{0.40} & \cellcolor{worst1}{0.3226} & 0.172 \\
 & 1.0 & \cellcolor{worst1}{0.28} & \cellcolor{best1}{0.43} & 0.47 & 0.3947 & \cellcolor{worst1}{0.153} \\
\midrule
\multirow{5}{*}{A4} & 0.1 & 0.48 & 0.47 & 0.47 & 0.4714 & 0.132 \\
 & 0.3 & \cellcolor{best1}{0.56} & 0.47 & 0.49 & \cellcolor{best1}{0.5056} & 0.068 \\
 & 0.5 & 0.51 & \cellcolor{best1}{0.47} & 0.48 & 0.4884 & \cellcolor{worst1}{-0.009} \\
 & 0.7 & 0.50 & \cellcolor{worst1}{0.35} & \cellcolor{worst1}{0.44} & 0.4311 & 0.072 \\
 & 1.0 & \cellcolor{worst1}{0.31} & 0.38 & \cellcolor{best1}{0.58} & \cellcolor{worst1}{0.4261} & \cellcolor{best1}{0.245} \\
\bottomrule
\end{tabular}
\end{table*}

\subsection{Prompt Sensitivity Ablation}
\label{app:prompt-sensitivity}

To verify that our metrics capture meaningful agent design differences rather than surface-level prompt variation, we conducted a prompt sensitivity ablation on the A3 (Demographic) specification. We created two paraphrased variants by rewriting all three components---role-playing framing, demographic formatting, and task instructions---while preserving informational content (Table~\ref{tab:prompt-sensitivity-design}). We measured within-paraphrase PAS variation (max--min range across the three A3 variants) on Study~2 ($N=256$) and Study~8 ($N=468$), selected for their moderate sample sizes (Table~\ref{tab:prompt-sensitivity-results}).

\begin{table}[!h]\small
\centering
\caption{\textbf{Prompt paraphrase design.} Three A3 variants with equivalent information but different surface forms.}
\label{tab:prompt-sensitivity-design}
\setlength{\tabcolsep}{3pt}
\begin{tabular}{@{}lccc@{}}
\toprule
 & \textbf{Role frame} & \textbf{Demographics} & \textbf{Task instructions} \\
\midrule
A3 & ``You are participating\ldots'' & Bullet list & ``Follow the experimenter's\ldots'' \\
Para.\ A & ``Imagine you are a person\ldots'' & Inline & ``Respond to each task\ldots'' \\
Para.\ B & ``You have been recruited\ldots'' & Indented block & ``Answer every question\ldots'' \\
\bottomrule
\end{tabular}
\end{table}

\begin{table}[!h]\small
\centering
\caption{\textbf{Prompt sensitivity results.} PAS variation from paraphrasing is negligible ($\leq 8.8$ pp) compared to cross-agent variation ($>32$ pp).}
\label{tab:prompt-sensitivity-results}
\begin{tabular}{@{}llc@{}}
\toprule
\textbf{Model} & \textbf{Study} & \textbf{PAS variation (pp)} \\
\midrule
Mistral Nemo & Study 2 ($N{=}256$) & 2.5 \\
Mistral Nemo & Study 8 ($N{=}468$) & 1.9 \\
DeepSeek V3.2 & Study 2 ($N{=}256$) & 8.8 \\
DeepSeek V3.2 & Study 8 ($N{=}468$) & 4.1 \\
\bottomrule
\end{tabular}
\end{table}

Across all four model--study combinations, PAS variation from paraphrasing is consistently small ($\leq 8.8$ percentage points), while cross-agent variation (A1--A4) exceeds 32 percentage points. This confirms that inference-level conclusions are robust to prompt surface form, and that the benchmark reliably distinguishes meaningful agent design differences from superficial wording changes.




\section{Broader Impacts}
\phantomsection\label{sec:impact}%
\subsection{Community Impact}
\label{commnity}

\textsc{HumanStudy-Bench} is designed to turn the accumulated 
archive of human-subject research into a living, reusable 
resource that the broader research community can reproduce, 
extend, and build on. We highlight three dimensions of impact.

\textbf{Open platform for community growth.}
\textsc{HumanStudy-Bench} is released as an open platform 
that treats participant simulation as an agent design problem. 
Practitioners can contribute validated published studies to 
expand the benchmark, design new experiments to pilot 
research hypotheses, evaluate and compare agent designs 
before deployment, and systematically diagnose where gaps 
to human behavior remain across diverse experimental 
paradigms.

\textbf{Preserving and operationalizing published experiments.}
The platform reconstructs full experimental protocols from 
original papers---stimuli, conditions, and statistical 
procedures---and instantiates them in a reusable execution 
engine, making decades of human-subject research reproducible 
in simulation without repeated manual effort.

\textbf{Rigorous inference-level evaluation.}
By applying the original statistical pipeline to both human 
and agent data, \textsc{HumanStudy-Bench} evaluates alignment 
at the level of inferential conclusions rather than surface 
outputs, aggregating heterogeneous evidence into comparable 
alignment scores via PAS and ECS.

\subsection{Ethical Considerations}
\label{ethic}

\textbf{Demographic representativeness.}
The benchmark's ground truth is anchored by 12 original 
studies plus 36 corroborating replications (3 per 
hypothesis). The originals span 
1946--2007 and were conducted on uniformly WEIRD (Western, 
educated, industrialized, rich, democratic) samples---chiefly 
US, UK, Canadian, and Dutch university students, with one 
US public-park sample~\citep{article} as the only
non-student exception. The replication suite extends the 
temporal range to 2018 and includes limited cross-cultural 
coverage (notably Henrich et al.~\cite{henrich2001search} on small-scale
non-WEIRD societies for the Ultimatum/Dictator hypothesis), 
but remains predominantly Western and educated. We therefore 
caution against interpreting benchmark alignment as universal 
human alignment, and encourage future extensions to 
incorporate broader cross-cultural replications as ground 
truth.

\textbf{Responsible use of published materials.}
All experimental materials used in \textsc{HumanStudy-Bench} 
are derived from published, peer-reviewed studies and are 
used solely for non-commercial research purposes. No new 
human subjects were recruited, and no personally 
identifiable information is collected or stored.

\textbf{Transparency and reproducibility.}
All pipeline outputs are human-verified before deployment, 
and all statistical analyses are executed deterministically 
using standard scientific libraries, ensuring that 
evaluation results are fully reproducible and independent 
of LLM stochasticity.

\end{document}